\documentclass[OptBiber]{jtcam_preprint}

\usepackage{biblatex}
\usepackage{verbatim}		
\usepackage{marvosym}		
\usepackage{tikzsymbols}	
\usepackage{tikz}			
\usepackage{algorithm}

\DeclareMathOperator*{\argmax}{arg\,max}
\DeclareMathOperator*{\argmin}{arg\,min}

\newtheorem{Modeling}{Modeling}[section]

\newtheorem{remarque}{Remark}[section]

\newtheorem{algo}{Algorithm}[section]
\newtheorem{procedure}{Procedure}[section]

\usepackage[normalem]{ulem}

\DeclareRobustCommand{\PhaseOne}{[\tikz[baseline=-\the\dimexpr\fontdimen22\textfont2\relax,inner sep=0pt,thick]\draw[red](0,0)--(.5,0);]}
\DeclareRobustCommand{\PhaseTwo}{[\tikz[baseline=-\the\dimexpr\fontdimen22\textfont2\relax,inner sep=0pt,thick]\draw[blue](0,0)--(.5,0);]}

\title{A probabilistic model for fast-to-evaluate 2D crack path prediction in heterogeneous materials}
\runningtitle{A probabilistic model for fast-to-evaluate 2D crack path prediction in heterogeneous materials}

\author[1,4]{Kathleen Pele}
\author[3,4]{Jean Baccou}
\author[2,4]{Loïc Daridon}
\author[1]{Jacques Liandrat}
\author[1]{Thibaut Le Gouic}
\author[2,4]{Yann Monerie}
\author[3,4]{Frédéric Péralès}
\runningauthor{K.Pele et al.}
\affil[1]{Aix Marseille Univ., CNRS, Centrale Marseille, I2M, UMR7353, 13451 Marseille, France  } 
\affil[2]{LMGC, Univ. Montpellier, CNRS, Montpellier, France} 
\affil[3]{Institut de Radioprotection et de Sureté Nucléaire (IRSN), Cadarache, France} 
\affil[4]{MIST, Univ. Montpellier, CNRS, IRSN, France}

\keywords{Markov chain, concrete, cracking prediction, machine learning}

\begin{document}
\nocite{*} 	
\begin{abstract}
This paper is devoted to the construction of a new fast-to-evaluate model for the prediction of 2D crack paths in concrete-like microstructures. The model generates piecewise linear cracks paths with segmentation points selected using a Markov chain model. The Markov chain kernel involves local indicators of mechanical interest and its parameters are learnt from numerical full-field 2D simulations of cracking using a cohesive-volumetric finite element solver called XPER.
This model does not include any mechanical elements. It is the database, derived from the XPER crack, that contains the mechanical information and optimises the probabilistic model.

The resulting model exhibits a drastic improvement of CPU time in comparison to simulations from XPER.


\end{abstract}


\section{Introduction}
Aging of materials is a major issue in industrial applications.  It is particularly the case in nuclear framework where the aging of cementitious materials can cause safety issues. Among various degradations due to aging, a special attention is devoted to concrete cracking 
induced by Internal Sulfate Attack (\cite{consequence_fissu}). These degradations generally lead to the development of a network of cracks.  
They considerably influence the strength of structures, reduce their tightness. Knowing the characteristics of cracking is essential in the study of the life extension of nuclear power plants.
\\
 At mesoscopic scale, concrete, used in  design of nuclear power plant containments, can be viewed as a two-phase composite material with mortar matrix and aggregates inclusions. The granular particles are randomly distributed in a matrix of mortar. The heterogeneities of this kind of material (\cite{quasifragile}) and the different associated scales (\cite{eche}) increase the complexity of its study.

In the context of research on the safety of Pressurized Water Reactors, IRSN, in collaboration with the LMGC (Mechanics and Civil Engineering Laboratory) through the joint laboratory MIST (Micro-mechanics and Integrity of Structure Laboratory), has developed a micro-mechanical approach for the analysis of the behavior of materials during hypothetical accidental transients (\cite{perales_th}). This approach is based on the concepts of Cohesive Friction Zone Model (MZCF) (\cite{MCZF}) associated with numerical modeling methods for multi-body systems based on the Non Smooth Contact Dynamics (NSCD) approach (\cite{NSCD}). The developed parallelized numerical platform, XPER (eXtended cohesive zone models and PERiodic homogenization) (\cite{perales,Xper2}), allows to simulate the initiation and two-dimensional propagation of multi-cracks in heterogeneous materials.
However, each simulation is very expensive in terms of computing time, since it can involve several days of computation on a few dozen processors. This CPU time can therefore become prohibitive in the context of probabilistic safety studies. 

In order to reduce this CPU time, this paper proposes the construction of a probabilistic model that allows to quickly predict a set of crack paths from the discretization of the microstructure associated with a local law of probability based on a Markov chain model. The transition core of the introduced model depends on two geometrical indicators. 
The parameters of the model are estimated from a set of training crack paths obtained numerically using full-field cohesive-volumetric finite element analysis with XPER. 
All the mechanics are contained exclusively in the training of the model.

 Several recent works are dealing with the development of statistical and machine learning tools for cracking data analysis. They study for example crack classification or detection of type of cracks (\cite{das,kim}). The crack path prediction has been also addressed in \cite{willot,corr_image} or \cite{bayar} but none of them leads to a surrogate model providing local information on the crack path  which is the originality of our development.
\\ 
This paper is organized as follows. The Section \ref{mech_analysis} is devoted to the reminder of some elements related to cracking of materials that are exploited in the construction of the model and allows to define working hypotheses. The Section \ref{sec_mech} describes the test case considered in this document and the computer code used for the simulations. The Sections \ref{indicators} and \ref{new_local} deal with the construction of the prediction model. After the introduction of the discretization of the problem, the two geometric indicators chosen to capture the local configuration of the aggregates are defined and an efficient procedure is proposed to evaluate them. On the basis of these indicators, a Markov chain model is then developed to perform the prediction of the crack path. Finally, the performance of the new model is studied in Section \ref{sec_appl} for different shapes of aggregates.

\section{Concrete crack}
\subsection{Phenomenology of cracking}
\label{mech_analysis}

\begin{figure}[htp!]
\center
\includegraphics[scale=0.5]{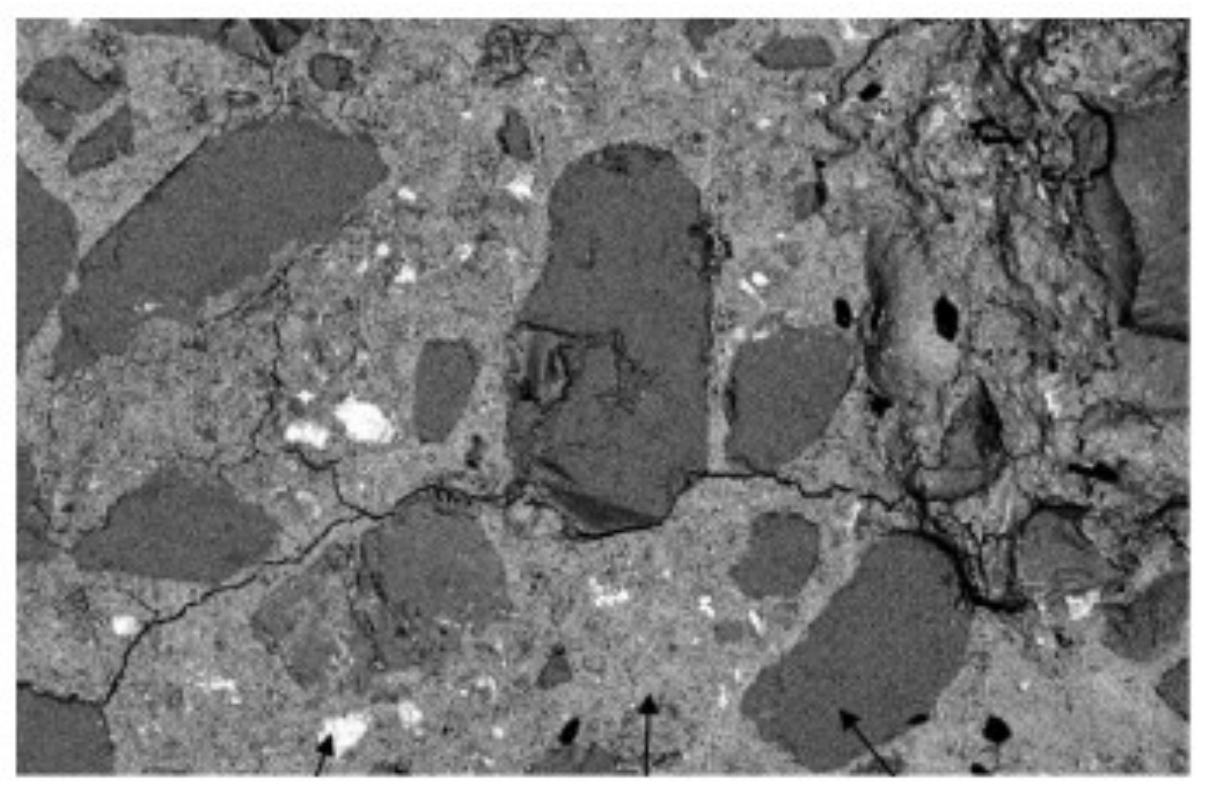} 
\caption{Example of a crack that propagates and follows the aggregate-matrix interface}
\label{fig:bypass}
\end{figure}

The objective here is to describe the behaviour of cracks in concrete in order to retain general assumptions that characterize the crack path. The crack path is directly related to the heterogeneous composition of the concrete. In general, cracking studies are performed at the mesoscopic scale where concrete can be considered as a bi-material composed of a matrix (mortar) and inclusions (aggregates). 
In this paper, rectangular microstructures with 25\% uniformly distributed aggregates of different shapes are studied.
Concrete is often considered as \textbf{quasi-brittle material} (\cite{quasifragile}).
It is clearly established that aggregates have a strong influence (\cite{aggreg_influ}) on the fracture faces. This influence is due, on the one hand, to a
\textbf{high fracture resistance of the aggregates} and, on the other hand, to \textbf{the relative weakness of the aggregate/matrix interfaces}. The interface properties depend on a zone around the aggregates called the Interfacial Transition Zone (ITZ) ( \cite{elice,Pope}). This area is very porous, reducing significantly its strength. Thus, the cracks, during their propagation, preferentially follow the aggregate/matrix interfaces (see \figurename~\ref{fig:bypass}) (\cite{aggreg_influ,interphase,interphase2})
.

It is established that aggregates and their boundaries have a significant influence on the mechanical behavior of concrete. Indeed, according to \cite{hussem}, the compressive strength
of a concrete depends for 21\% on cement paste, for 12\% on aggregates and 67\% on interphase. 
To summarize, the presence of aggregates leads both to (\cite{WB}): 
\begin{itemize}
\item a heterogeneity of local mechanical fields, 
\item an increase in areas of weakness
(aggregate/matrix interfaces) which locally constitute privileged cracking zones.
\end{itemize}

In addition, the local stress state implies \textbf{a direction of propagation of the crack}. The crack displacements are therefore restricted by a half-plane oriented ahead of the crack tip due to the imposssibility to move backwards.
\\
This analysis of the behavior of cracking in concrete allows to identify four working hypotheses:
\begin{itemize}
\item[(H1)] the matrix is elastic brittle,
\item[(H2)] the aggregates are stiff and unbreakable,
\item[(H3)] the matrix aggregate interface is the weakest zone in the composite material,
\item[(H4)] the crack propagates mainly in mode I and thus in a half-plane oriented in the local direction of propagation, in front of the crack tip perpendicularly to the local tensile stress
\end{itemize}
\subsection{Numerical simulation of cracking }
\label{sec_mech}
\label{lim_mech}

\begin{figure}[h!]
\center
\includegraphics[scale=0.09]{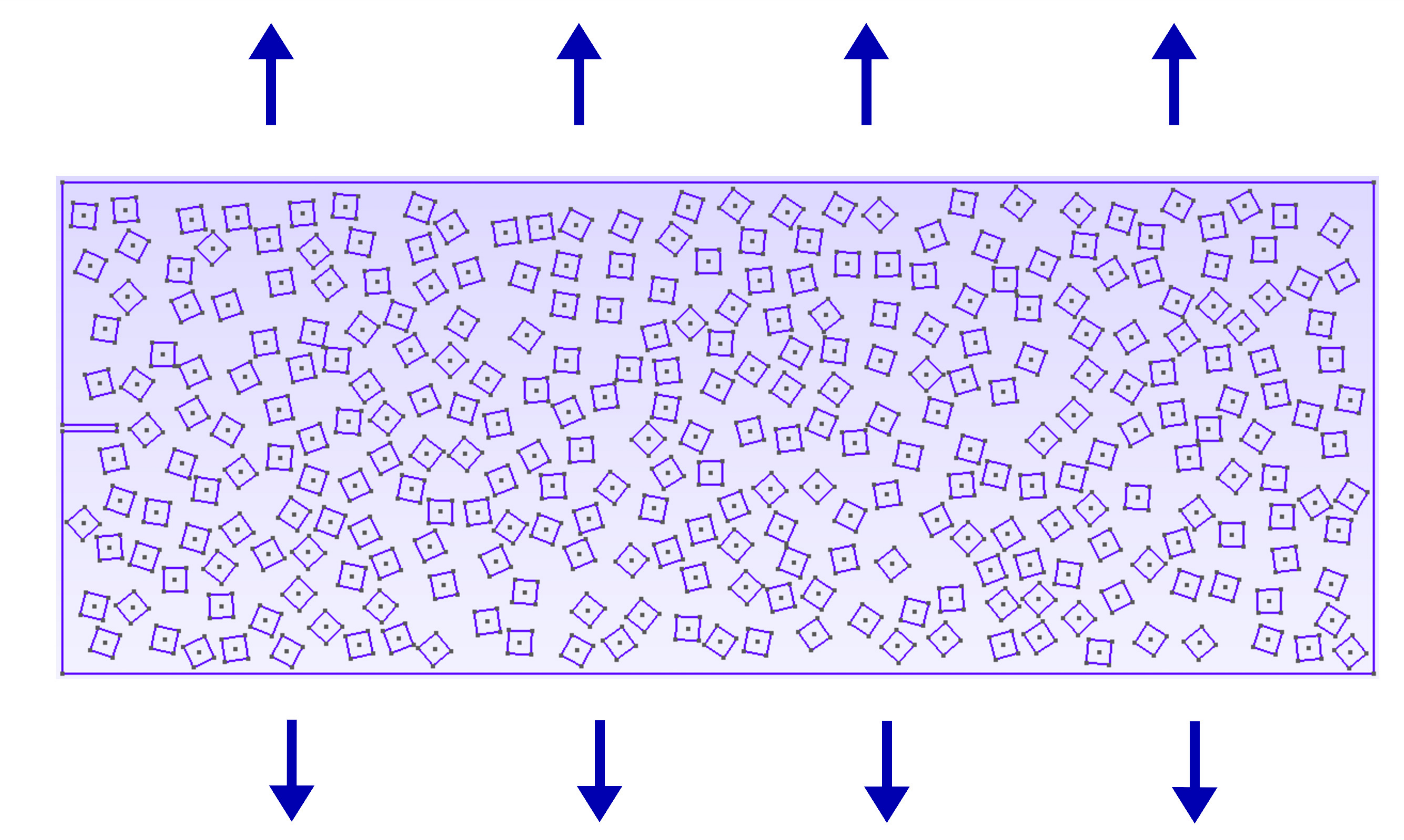} 
\begin{center}
\caption{\label{fig:testcase}Test case submitted to traction (drawing process for the distribution of aggregates in the microstructure : HCPP (Hard Core Point Process)) }
\end{center}
\end{figure}

In this section, the numerical representations of the concrete and the XPER software, that allows to numerically simulate the crack, will be presented.
Numerical samples of concrete are constructed from morphological descriptors.
XPER implementation involves a multi-contact modeling
strategy based on the Non-Smooth Contact Dynamics (NSCD) (\cite{NSCD}) method where cohesive models are introduced as mixed boundary conditions between each volumetric finite element. Each element or group of elements of the mesh can be considered as an independent body and the interface between bodies follows a frictional CZM (\cite{czm2}) with no regularization nor penalization (\cite{perales}).
 These descriptors are statistical and geometric informations identified on real concrete, like n-point moments and spatial covariance (\cite{jeulin}) allowing to characterize the spatial arrangements of different phases. The numerical microstructure are qualified as statistically similar to real concrete in the sense of these descriptors. An example of such microstructure is given in  \figurename~\ref{fig:testcase}.

The test case deals with a rectangular domain ($l \times L = 0.225m \times 0.600m$) submitted to a symmetrical traction on both sides (up and down) of the sample (\figurename~\ref{fig:testcase}).
The microstructure, composed of a matrix with 25\% of polygonal aggregates (square shapes in the \figurename~\ref{fig:testcase}), is considered as isotropic. We postpone to Remark \ref{vf_comment} for comments on the choice of this volume fraction. Cracks can propagate through the matrix and not through aggregates (assumptions (H1) and (H2)).

\noindent All full-field crack simulations are performed with XPER code. The CPU time for a complete microstructure cracking simulation like \figurename~\ref{fig:testcase} is about 44 hours for a computer : Intel(R) Xeon(R) Gold 6126 CPU @ 2.60GHz connected in Infiniband - 256Go RAM - Nodes of 24 processors. 
The simulation parameters are given in \tablename~\ref{tab1-1} to \tablename~\ref{tab1-3}. 
For more information on the  parameters, we refer the reader to the following article \cite{para_table2} and the following theses \cite{bichet}, \cite{soc}. These parameters are chosen in the case where there are no associated experimental measurements.

\begin{table}[!h]
\begin{center}
\begin{tabular}{|l|c|c|}
  \hline
 & Matrix & Aggregates \\
  \hline
 $\rho$ (kg/$m^3$) & 2900 & 2900  \\
 E (Pa) & 12e9 & 60e9 \\
 $\nu$ & 0.2 & 0.2 \\
  \hline
\end{tabular}
\caption{Volumic parameters of the simulations}
\label{tab1-1}
\end{center}

\end{table}

\begin{table}[!h]

\begin{center}
\begin{tabular}{|c|c|c|c|}
  \hline
  & Matrix-Matrix & Aggregate-Aggregate & Matrix-Aggregate \\
  \hline
  $C_n$=$C_t$ (Pa/m) & 1e17 & 1e17 & 1e17 \\
  $\sigma_0$ (Pa) & 4.6e7 & 2.4e9 & 1.4e7 \\
  w (J/$m^2$) & 20 & $\infty$ & 20 \\ 
  \hline
\end{tabular}
\caption{Surface parameters of the simulations }
\label{tab1-2}
\end{center}

\end{table}

\begin{table}[!h]

\begin{center}
\begin{tabular}{|c|c|}
\hline
Area of aggregates & 0.6$\times$0.225 $m^2$ \\
Degrees of freedom & 48666 \\
Number of processors & 24 \\
Computation time & 44h \\
\hline
\end{tabular}
\caption{Numerical parameters of the simulations}
\label{tab1-3}
\end{center}

\end{table}

\begin{figure}[h!]
\center
\includegraphics[scale=0.2]{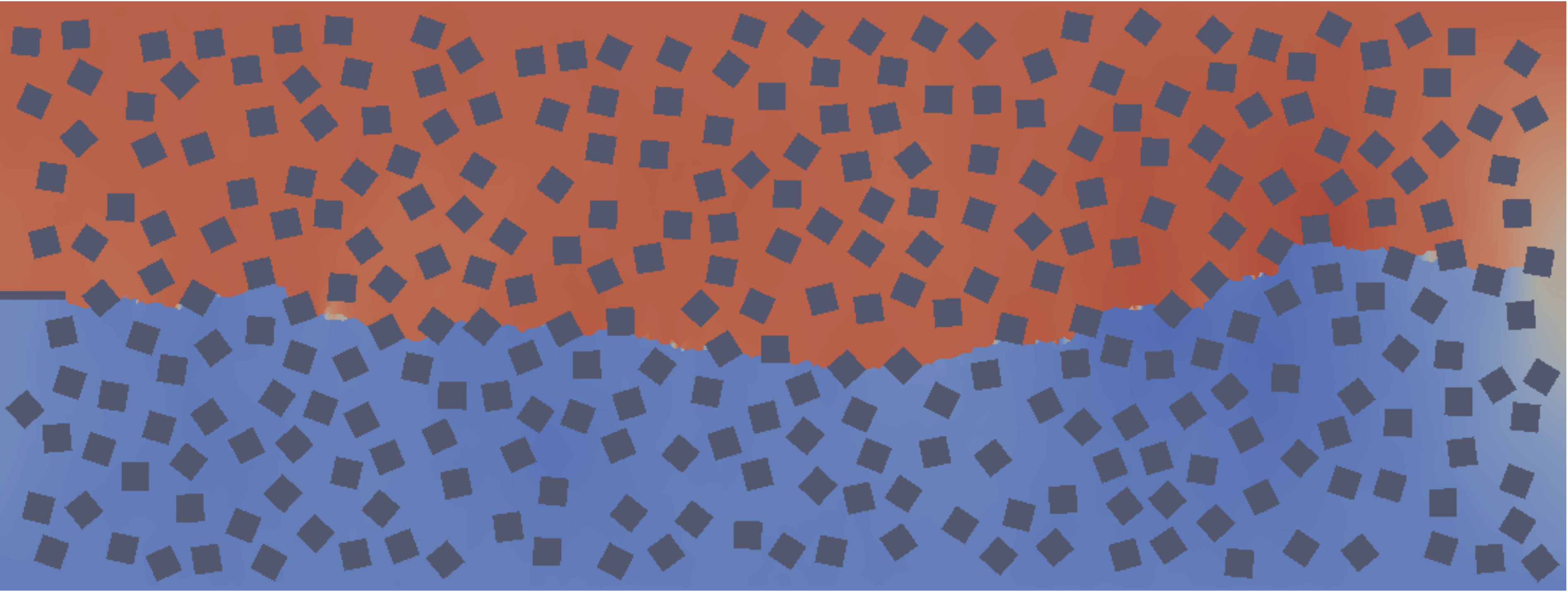} 
\caption{\label{fig:XPERconcrete} Crack path (colour discontinuity) obtained with XPER for the case described in the \figurename~\ref{fig:testcase}. The colours stand for the vertical displacement.}
\end{figure}

\noindent \figurename~\ref{fig:XPERconcrete} shows an example of  simulation with square-shape aggregates. 

The crack is represented by the colour discontinuity (colour stand for vertical displacement). The main direction of crack path is therefore almost perpendicular to the direction of the load.
\\
It turns out that two numerical concretes sharing the same morphological descriptors (covariogram, percentage of aggregate) can exhibit different crack paths. 

\begin{figure}[h!]
\center
\includegraphics[scale=0.4]{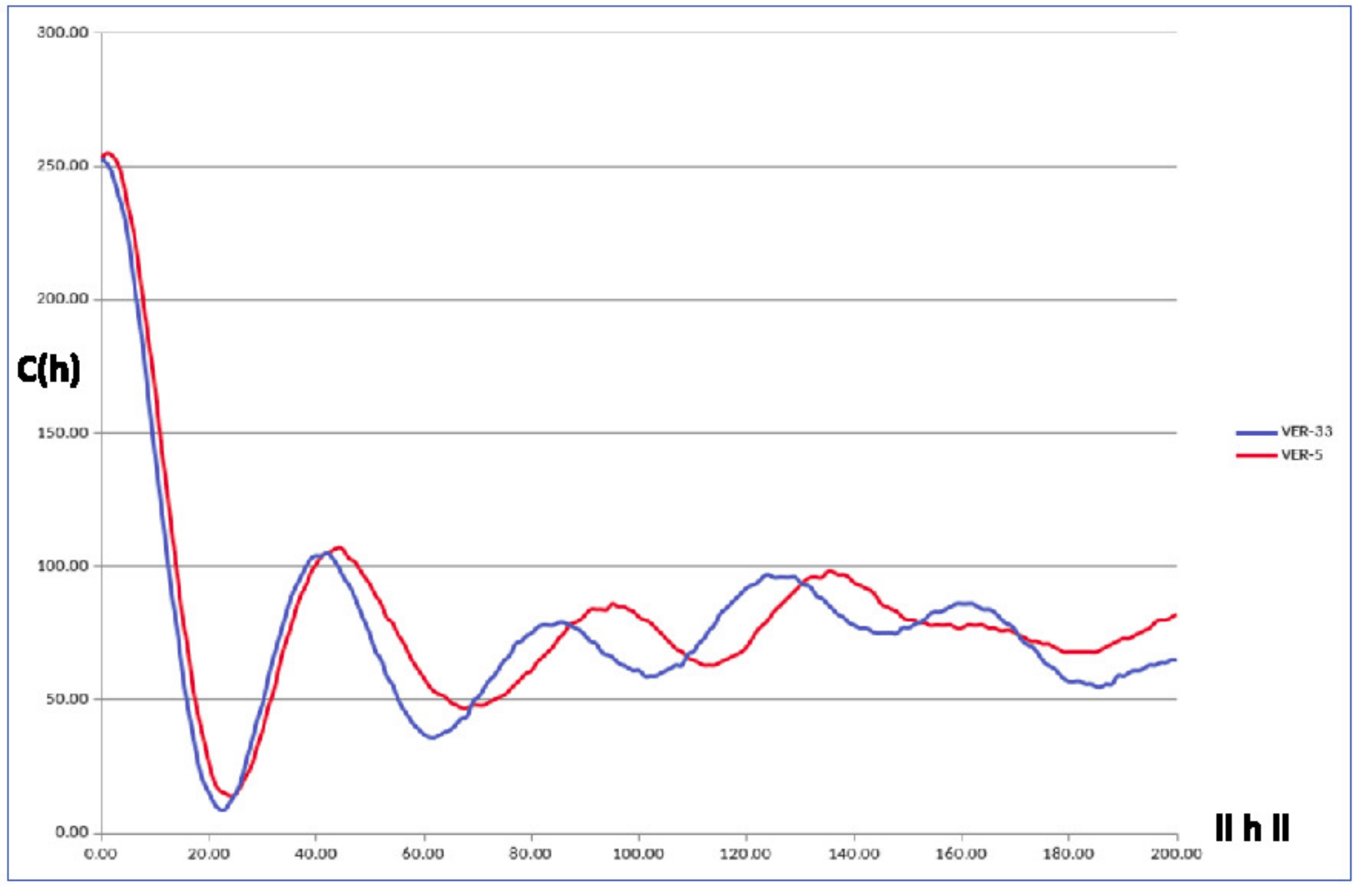} 
\caption{\label{fig:compcov} Example of covariograms for two differents microstructures and associated cracking faces. 
$C(\|h\|)$ represents the value of the covariogram and $\|h\|$ is the norm of the translation vector}
\end{figure}

To illustrate this point, we propose an example on the \figurename~\ref{fig:compcov} with the covariogram (\cite{jeulin}) under stationary and isotropic assumptions.
The covariogram allows to study the spatial distribution and the relative organisation of the phases of a random medium. Mathematically, it represents the probability that for a point x located in a given phase $A$, here an aggregate, $x + h$, where $h$ is a translation vector, is in the same phase:
\[C(x, x + h) = P (x \in A, x + h \in A)\]
In the case of a stationary and isotropic process, the covariogram only depends on norm of the translation vector $h$ and is noted for simplicity $C(\|h\|)$.

\begin{figure}[h!]
\center
\includegraphics[scale=0.3]{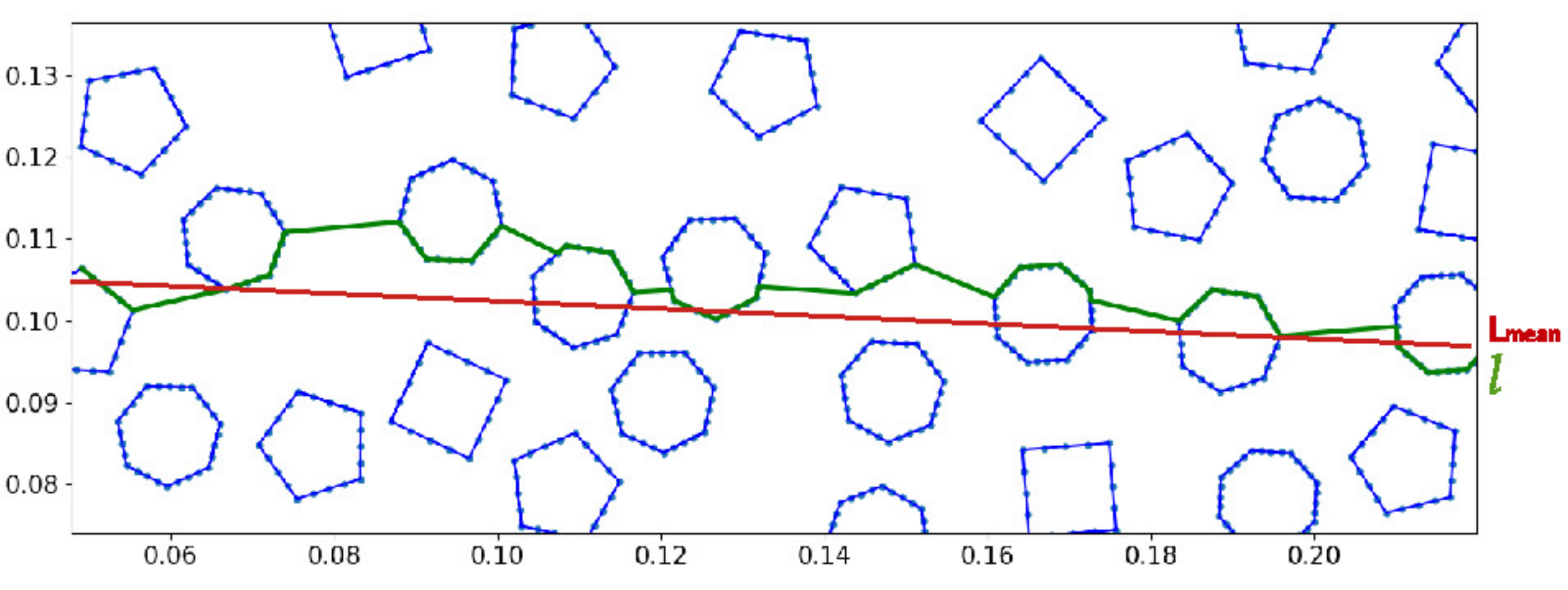} 
\caption{\label{fig:tortu} Representation of the tortuosity : $\tau=\frac{l}{L_{mean}}$}
\end{figure}

As a result, they lead to different values for mechanical quantities of interest such as the tortuosity (\figurename~\ref{fig:tortu}), which definition is  recalled below.
\begin{definition}[Tortuosity]: the tortuosity $\tau$ of a crack is the ratio between the length $l$ of the crack path and $L_{mean}$ the length of the line joining the first and last points of the crack: $ \tau =\frac{l}{L_{mean}}$.
\end{definition}

 In view of the cracked microstructure of \figurename~\ref{fig:XPERconcrete}, it appears that the \emph{local} configuration of the aggregates strongly influences the path of the crack. This information is central in our construction of the new crack indicators.


 

\section{Definition and evaluation of local crack indicators}
\label{indicators}
In this section, several key notions are presented in order to satisfy the hypotheses (H1) to (H4) (Section \ref{mech_analysis}) describing the local behavior of a crack. The definition and the evaluation of the local crack indicators are described in subsections \ref{sec:indic} and \ref{sec:eval}. 

\subsection{Discretization  of the microstructure}
\label{sec_not}
\begin{figure}[h]
 \center
 \includegraphics[scale=0.5]{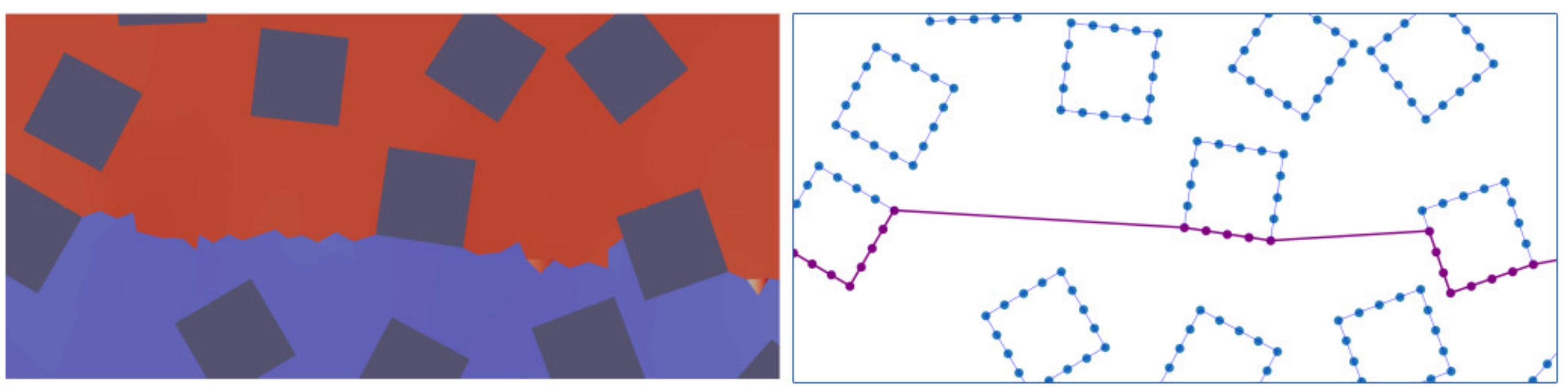} 
 \caption{\label{fig:discre} Discrete representation of a microstructure and   crack path: on the right, the blue dots stand for the discretization points $(y_i^E)$ and the purple path represents the XPER crack crossing the points $(x_i^E)$.}
\end{figure}

Although the main crack propagates in the mode I direction, the local propagation of a crack depends strongly on the microstructure configuration (the aggregates) near the crack tip. Therefore, we propose a discretization of the microstructure, at this scale, allowing to estimate the step by step crack path. 
 In this work, each aggregate is approximated by a polygon where each side is discretized by 5 points.
This allows to simulate the propagation of the crack along the side (\figurename~\ref{fig:discre}).
The crack path can leave or join an aggregate at several positions along the side of the aggregate and not only at the corners of the polygon.

  The matrix is not discretized which implies that either the crack crosses the matrix following a straight line between two points of the discretization belonging to two distinct aggregates (assumption (H1) of Section \ref{mech_analysis}) or it follows the side of an aggregate (assumption (H3) of Section \ref{mech_analysis}). 
\\
\\
For a microstructure $E$, the set of discretization points denoted by $\mathcal{Y}_E=\{y^E_i\}_{i=1,\dots,N_E}$ constitutes the discrete granular microstructure (\figurename~\ref{fig:discre}). The crack path is then approximated from a subset of $\mathcal{Y}_E$ 
denoted by $\mathcal{X}_E=\{ x^E_i\}_{i=1,\dots,m_E}$. We can define the discretized crack as follows.
\begin{definition}[Discretized crack]: \label{def_fiss}if one considers the Cartesian coordinate system whose origin is at the bottom left corner of the microstructure, the crack path can be parametrized by 
$\left(x^1,h_E\left(x^1\right)\right)$ where $
h_E: \mathbb{R}_{+} \rightarrow \mathbb{R}$ is a piecewise linear function such that: if $\left(x_{i}^{E,1},x_{i}^{E,2}\right)$ denotes the coordinate of $x^E_{i}$ in the Cartesian system, 

\label{linear_app}
\begin{eqnarray*}
\forall x^1 \in \left[x_{i}^{E,1},x_{i+1}^{E,1} \right], \ h_E\left(x^1 \right)=\frac{x^{E,2}_{i+1}-x^{E,2}_{i}}{x^{E,1}_{i+1}-x^{E,1}_{i}} \left( x^1-x_{i}^{E,1}\right)+x^{E,2}_{i}.
\end{eqnarray*}
\end{definition}



\begin{figure}[!h]
 \center
 \includegraphics[scale=0.4]{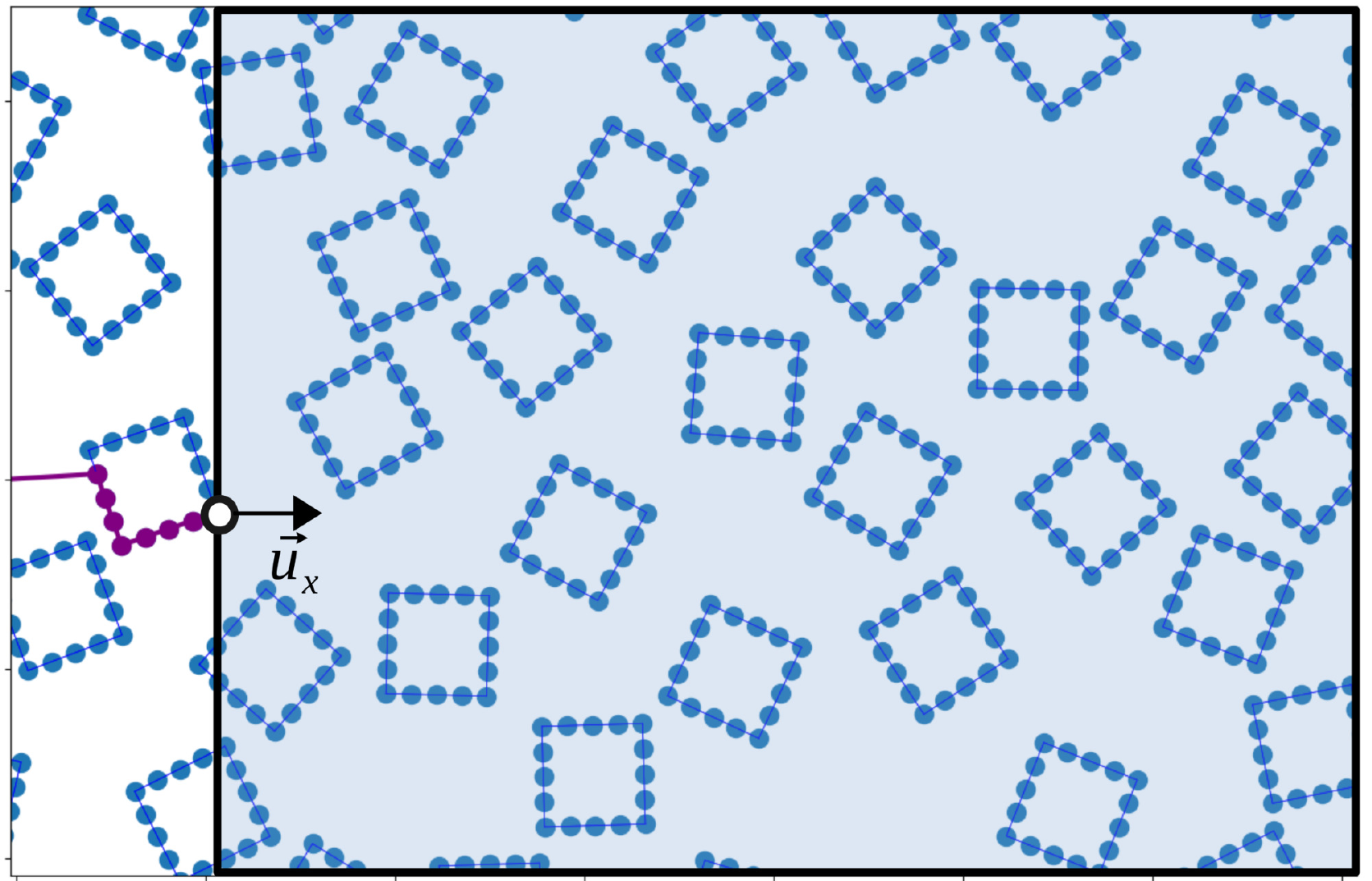}
 \caption{ Example of a \emph{field of view} (blue area) at the crack tip (white circle), following the local direction of propagation $u_x$. The purple line stands for the path of the crack }
  \label{fig_field}
 \end{figure}

Assumption (H4) implies that, starting from a point of the discretized crack path, only a subset of discretization points can be reached. They are located in a \emph{field of view} defined as follows (an example is exhibited in \figurename~\ref{fig_field}).

\begin{definition}[Field of view]:
\label{def_f}if $E$ is a microstructure, $x \in \mathcal{X}_E$ the crack tip and $\vec{u_x}$ is the local direction of crack propagation at $x$, the \emph{field of view} of the crack path at $x$ is defined by the area containing the set of $I$ discretization points $\{y^E_i\}_{i \in I} \subset \mathcal{Y}_E$ such that:
\begin{eqnarray}
\langle\overrightarrow{xy^E_i},\overrightarrow{u_x}\rangle\geq 0, \forall i \in [1,\hdots, N_E]
\end{eqnarray} 
 where $\langle \ , \ \rangle$ denotes the euclidean scalar product in $\mathbb{R}^2$.
\end{definition}

\begin{remarque}
Note that it is possible to take into account a non constant local direction and locally orientate the field of view. In the numerical studies of this paper, $\vec{u_x}$ is a constant vector orthogonal to the loading direction. We refer to chapter 4 of \cite{pele_th} for an example where $\vec{u_x}$ is evaluated through a mechanical simulation.     
\end{remarque}
Depending on the configuration associated of the tip of the crack, the field of view may not contain point in the same aggregate. 
This strongly influences the crack path and should be taken into account in the prediction model developped in this paper. Consequently, it is important to distinguish the two following configurations of the \emph{field of view}:
%
\begin{itemize}
\item \textbf{Configuration F1: } \emph{field of view} including possibilities to follow the aggregates sides or to cross the matrix (\figurename~\ref{fig:F1}),
 \begin{figure}[h!]
\center
\includegraphics[scale=0.4]{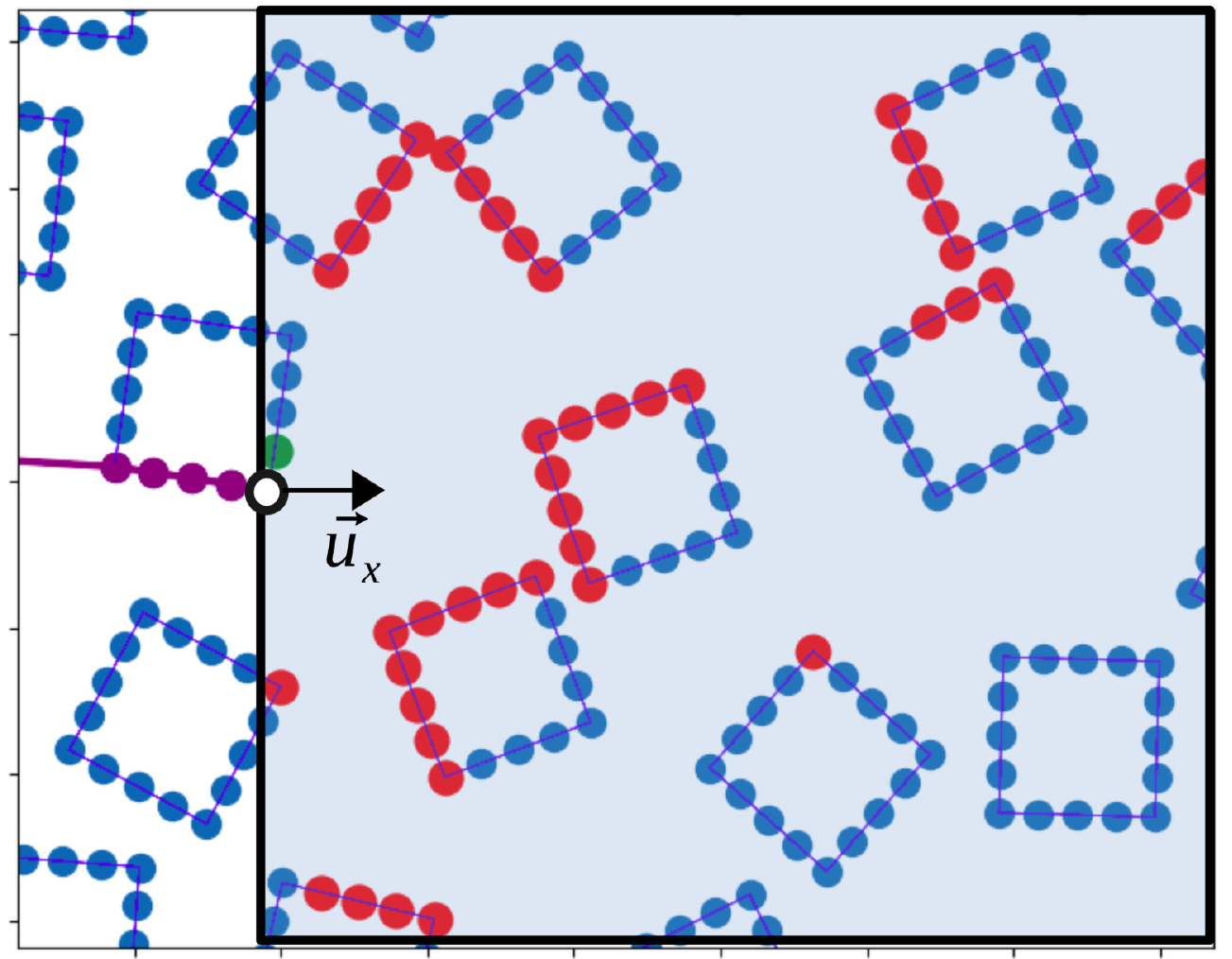} 
\caption{\label{fig:F1} Field of view (F1): the green point stands for the candidate point belonging to the original aggregate, the red points stands for the candidate points on the other aggregates (implying to cross the matrix). The purple line stands for the path of the crack}
\end{figure}
 \item \textbf{Configuration F2:}  \emph{field of view} only including possibilities to cross the matrix (\figurename~\ref{fig:F2}).
\begin{figure}[h!]
\center
\includegraphics[scale=0.4]{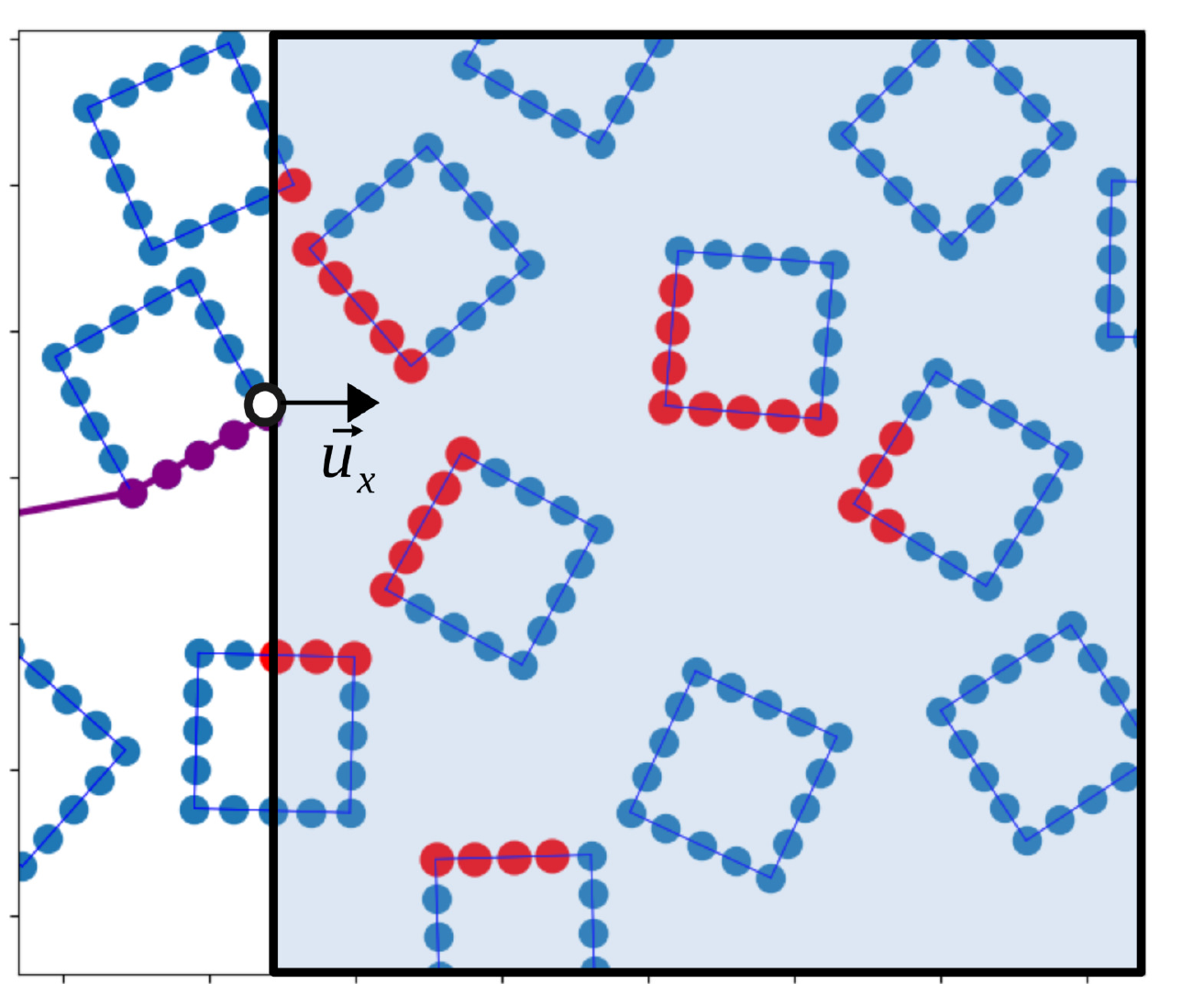} 
\caption{\label{fig:F2} Field of view (F2) : the red points represent the candidate points on the other aggregates (implying to cross the matrix). The purple line stands for the path of the crack.}
\end{figure}
\end{itemize}

\subsection{Definition of the two indicators}
\label{sec:indic}

This section is devoted to the definition of two geometrical indicators to capture the local behavior of a crack. These indicators will characterize all the couples $\left(x,y^E_i\right)$ where $x$ is the tip of the crack and $y_i^E$ a point of the field of view E. The first indicator is the angle that each vector $\overrightarrow{xy^E_i}$ makes with the local direction of propagation, $\overrightarrow{u_x}$  (assumption (H4)). The second indicator is the Euclidean norm of $\overrightarrow{xy_i^E}$. These indicators are evaluated at each increment of the crack propagation.

\begin{definition}[Crack indicators]:
\label{def_descr}let $E$ be a microstructure, $x\in \mathcal{X}_E$ be the tip of the crack and $\vec{u_x}$ be the local direction of propagation at $x$. If $\langle \ , \ \rangle$, resp. $||.||$, is the euclidean scalar product, resp. norm,  in $\mathbb{R}^2$, for any $y$ in the \emph{field of view} of $x$, the two local indicators are defined by:

\begin{eqnarray*}
d_{x}\left(y\right)&=& || \overrightarrow{xy}||,\\
\theta_{x}\left(y\right)&=&\arccos \left( \frac{||\langle \overrightarrow{xy},\overrightarrow{u_x} \rangle)||}{||\overrightarrow{xy}||\cdot  ||\overrightarrow{u_x}||}\right),
\end{eqnarray*}

For any $x$, normalized distance and angle are also introduced as:

\begin{eqnarray*}
\tilde d_{x}\left(y\right)&=&\frac{d_x(y)-d_{min}}{d_{max}-d_{min}},\\
\tilde \theta_{x}\left(y\right)&=&\frac{\theta_x(y)-\theta_{min}}{\theta_{max}-\theta_{min}},
\end{eqnarray*}

 where $d_{max},d_{min}$ and $\theta_{max},\theta_{min}$ are the maximum and minimum values of the indicators associated to the candidate points in the field of view for a given location of the crack tip.

 \end{definition}
 
\begin{figure}[h!]
    \centering
    \includegraphics[scale=0.35]{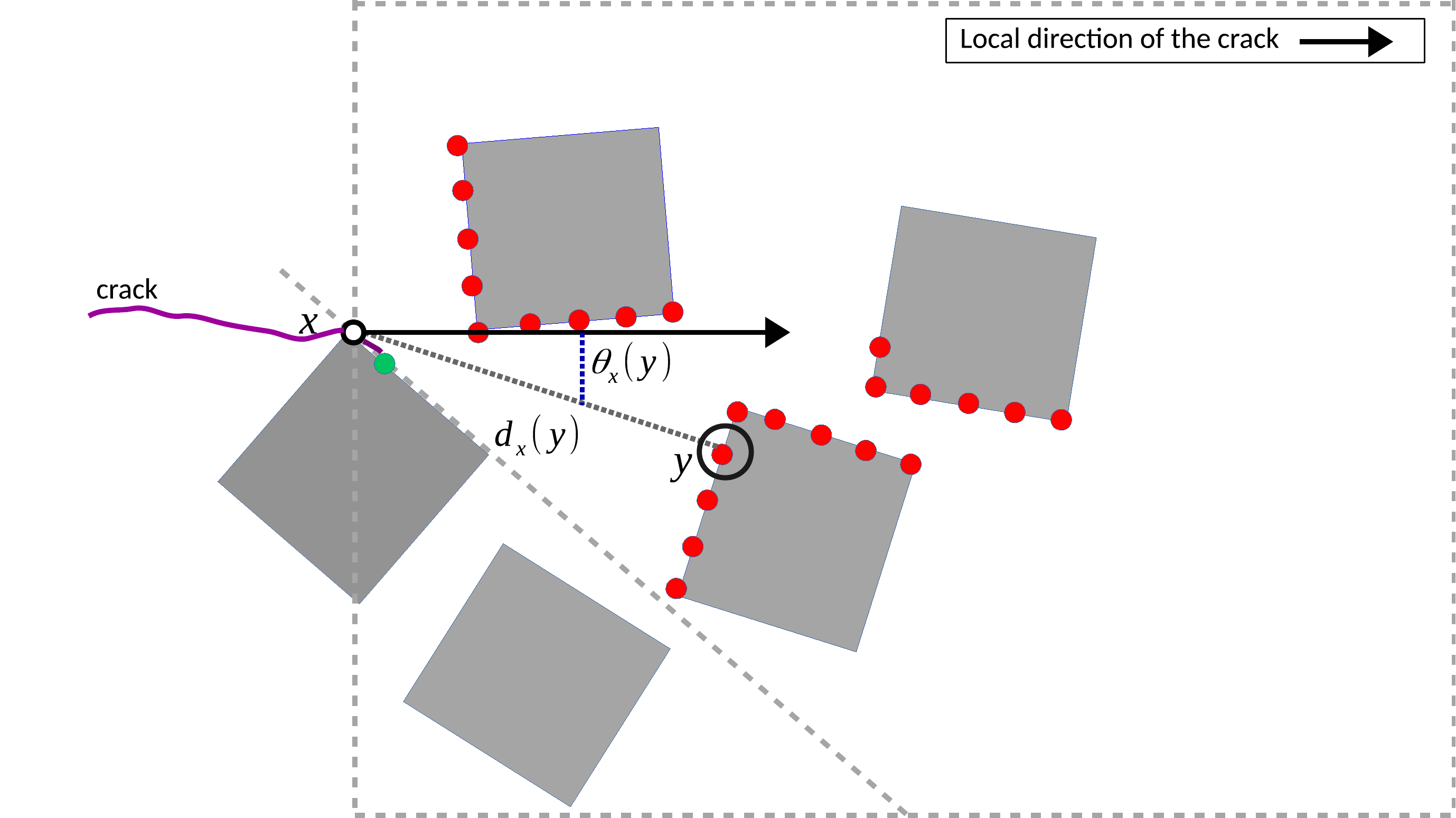}
    \caption{Illustration of the indicators of the candidate point $y$ for the crack tip $x$ in the configuration F1. The red points correspond to candidate points on neighbouring aggregates and the green point to the possibility of going
along the same aggregate. }
    \label{fig:descri}
\end{figure}

The \figurename~\ref{fig:descri} illustrates the defined indicators.

\subsection{Evaluation of the indicators} 
\label{sec:eval}
This section is devoted to the evaluation of the two previous indicators. 
In the first section \ref{proc_ind},  an efficient procedure to select the points actually reachable by the tip of the crack in the field of view is developped. In the second section \ref{sec_app_ind}, a numerical study of these indicators is performed by analyzing a training set of microstructures cracked with XPER code.

\subsubsection{Procedure to select the candidate points}
\label{proc_ind}
For each point $x$ of the crack, a straigthforward strategy for indicator evaluation would 
require the computation of the distance and the angle, $d_x\left(y_i^E \right)$ and $\theta_x\left(y_i^E\right)$, for all the discretization points located in the \emph{field of view} of $x$. However, following Assumption (H2), some  points cannot be reached by the crack due to the presence of aggregates and should be deleted from the candidate points. For a given aggregate, the deletion procedure is based on the detection of an area not visible from $x$ because it is hidden by the aggregate. This area is called in the sequel the shadow area.
 
\begin{figure}[h!]
    \centering
    \includegraphics[scale=0.3]{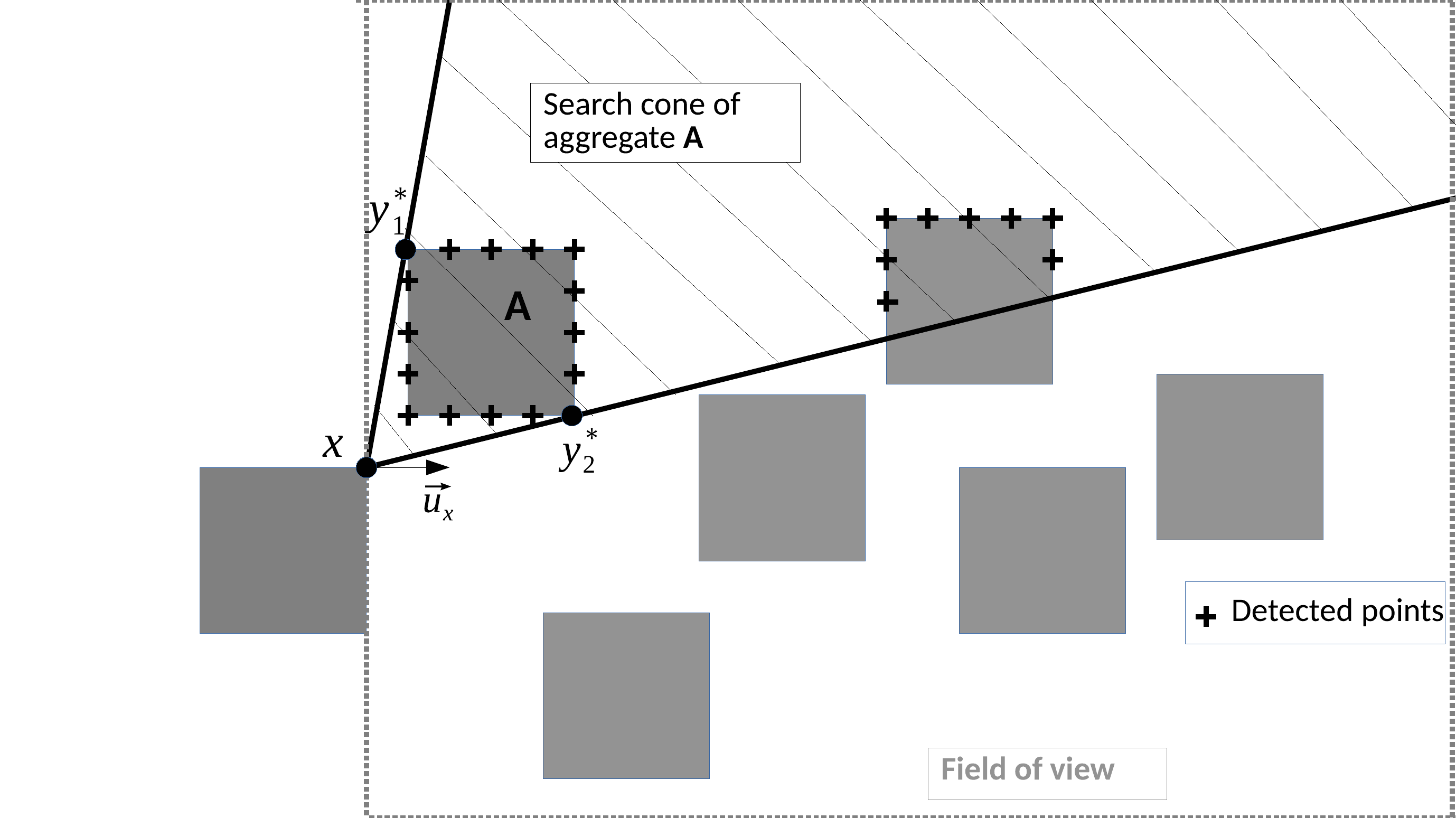}
    \caption{Illustration of Procedure 3.1 : construction of the search cone (hatched area) associated to aggregate A.}
    \label{fig:shadow}
\end{figure}

\begin{figure}[h!]
    \centering
    \includegraphics[scale=0.3]{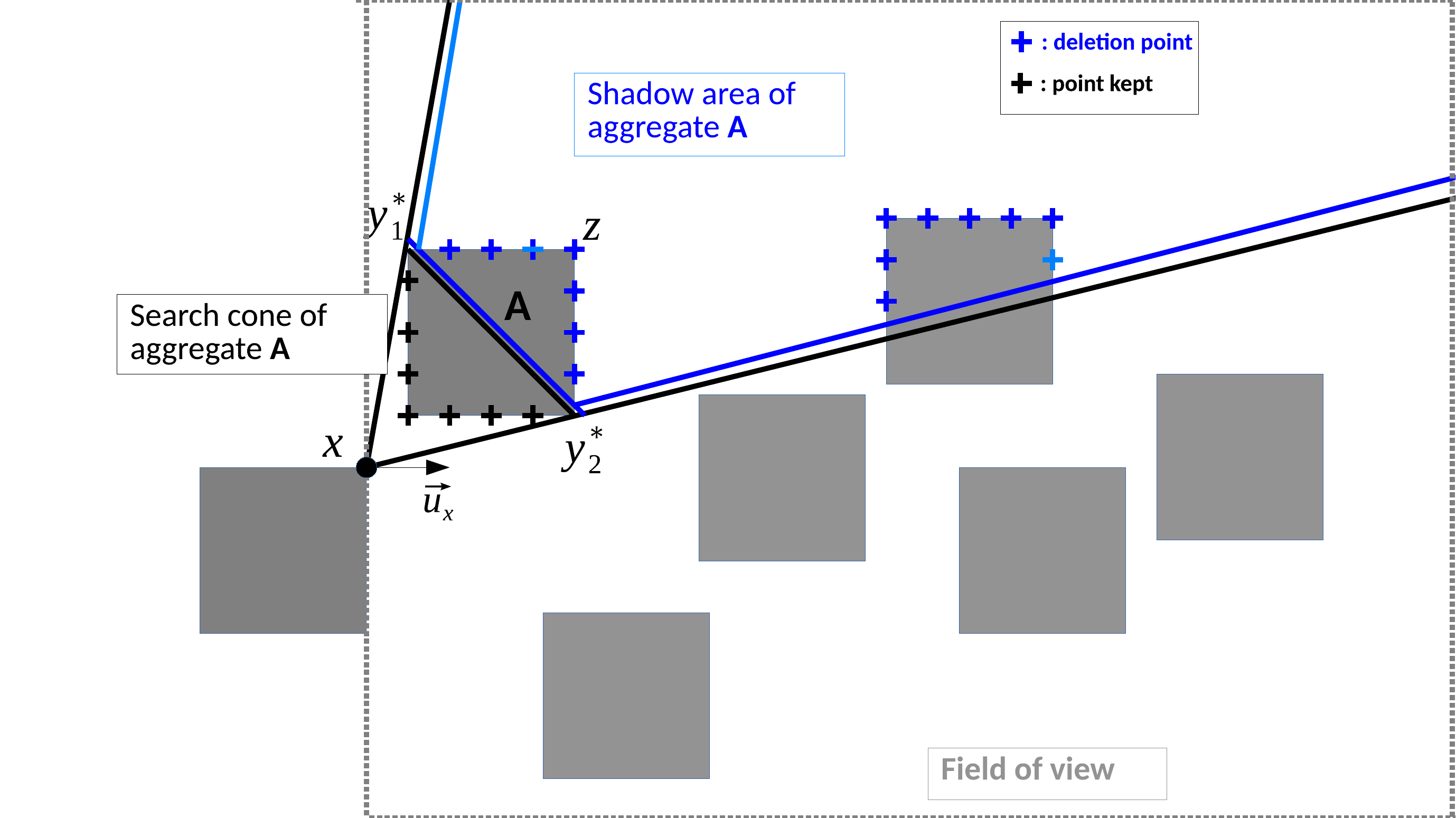}
    \caption{Illustration of Procedure 3.1 : construction of the shadow area (blue area) associated to aggregate A }
    \label{fig:area_vis}
\end{figure}

The method to identify this shadow area and to delete the corresponding discretization points is given by Procedure \ref{prop_penombre}. To prevent a dependence to the local crack propagation orientation, the position and the shape of the aggregate, it relies on computational geometry (\cite{geoalgo}).
\begin{procedure}[Deletion of the discretization points in the shadow area of an aggregate]:
\label{prop_penombre}let $E$ be a discrete granular microstructure, $x \in \mathcal{X}_E$ be a crack tip and $\{y_j^E\}_{j=1,\hdots,n_E}$ be the $n_E$ points discretizing the sides of an aggregate $A$ in the field of view of x. For any $(i,k) \in \{1,\hdots,n_E\}^2$, we denote by $\angle (\overrightarrow{x y_i^E},\overrightarrow{ x y_k^E})$ the angle between two vectors $\overrightarrow{x y^E_i}$ and $\overrightarrow{x y^E_k}$ in $[0,\pi]$.
The deletion is based on the identification of the points lying in the shadow area of the aggregate. We first find the two points $y_1^{*}$ and $y_2^{*}$ such that  
$\angle(\overrightarrow{ xy_1^{*}}xy_2^{*} \rangle = \max_{(i,k) \in \{1,\hdots,n_E\}^2} \angle (xy_i^{E}xy_k^{E}) $. 
This angle defines a search cone (\figurename~\ref{fig:shadow}).
A point $z$ in the field of
view belongs to the cone if : 
  
\begin{eqnarray*}
  det\left(\overrightarrow{xy_2^{*}},\overrightarrow{xz}\right)  det\left(\overrightarrow{xy_1^{*}},\overrightarrow{xz}\right)<0,
\end{eqnarray*}    
  

\noindent A point $z$ in the cone belongs to the shadow area and is deleted if (\figurename~\ref{fig:area_vis}):
\begin{eqnarray*}
det\left(\overrightarrow{xy_1},\overrightarrow{y_1^{*}z}\right)  det\left(\overrightarrow{y_1^{*}z},\overrightarrow{y_{1}^{*}y_{2}^{*}}\right)<0.
\end{eqnarray*}

\end{procedure}

\noindent This procedure is repeated for all aggregates, allowing the capture of the candidate points of the field of view  respecting the assumption ($H_2$).




\newpage
\subsubsection{Analysis of the relevance of the indicators}
\label{sec_app_ind}

\begin{figure}[h!]
\includegraphics[scale=0.45]{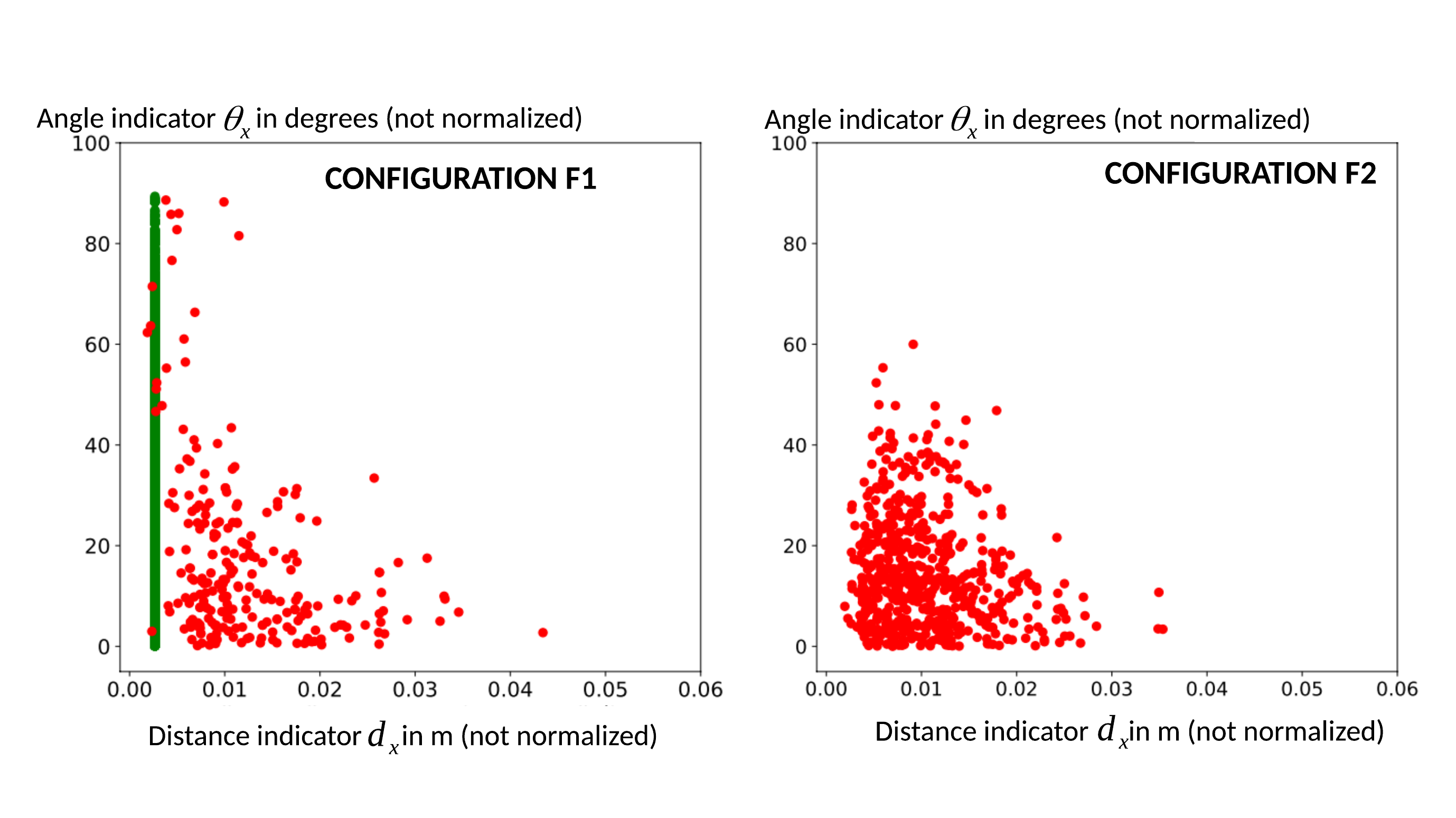}  
\caption{\label{fig:courbegran} Indicator values in configurations F1 and F2 associated to points on the crack for $35$ microstructures with square aggregates. Each point represents a couple of not normalized values $\left(d_x,\theta_x\right)$. The red dots stand for candidate points present on neighbouring aggregates (cross the matrix) and the green dots stand for the candidate points belonging to the same aggregate }
\end{figure}

In this section, we analyze the relevance of the two indicators angle and distance to capture the local characteristics of the crack path. This analysis is also exploited to define the different terms of the prediction model developped in section \ref{new_local}. \\
The two indicators are evaluated for each discretization point located on the crack path using Procedure \ref{prop_penombre} for a set of 35 microstructures whose cracking is simulated by the XPER code following Section \ref{lim_mech}. In this study the local direction of the field of view is always orthogonal to the loading (mode I). The indicator evaluation is performed for the two configurations of the field of view : F1 (see \figurename~\ref{fig:F1}) and F2 (see \figurename~\ref{fig:F2}). \figurename~\ref{fig:courbegran}  displays the results of the set of indicator pairs selected by the cracks of the 35 microstructures.


\begin{itemize}
    \item Configuration F1 (\figurename~\ref{fig:F1})
\end{itemize}
\figurename~\ref{fig:courbegran} (left-hand side graph) confirms that the crack is likely to follow the aggregate side. This situation corresponds to the large number of green points with the shortest distance and a variation range between $0$ and $\frac{\pi}{2}$ for the angle. For these points on the same aggregate, the distance indicator is not influential.
 It is interesting to observe that the green points are all located on a vertical line since the discretization of each aggregate is uniform. Therefore, when the crack is propagating along the aggregate side, the points reached by the cracks are equally spaced leading to the same
value for the distance indicator.
\\
It is important to note that the attraction of the point on the same aggregate may be reduced in some cases. Indeed, the observation of our training base highlights that the crack sometimes chooses to leave the aggregate side and to cross the matrix even if it has the possibility to follow the aggregate (red points on \figurename~\ref{fig:courbegran}). This change of behaviour in the crack propagation is due to two situations that simultaneously occur: following the same aggregate involves a large angle while crossing the matrix is associated to the combination of small angle and short distance.

\begin{itemize}
\item  Configuration F2 (\figurename~\ref{fig:F2}) 
\end{itemize}

In this configuration (right-hand side graph of \figurename~\ref{fig:courbegran}), since there is no candidate point on the aggregates sides, the crack can only propagate into the matrix. The indicators values of the candidates points reached by the crack are in the bottom left hand corner of the graph (i.e short distance and small angle). This reflects the fact that the crack is at the same time constrained by the direction of the stress and attracted by the closest aggregates.

\begin{remark}
\label{vf_comment}
The previous numerical tests show that the crack can reach an aggregate further way from the crack tip but located in the direction of propagation. This situation does not occur for high volume fraction of aggregates. That is why a volume fraction of 25\% has been chosen for the analysis provided in this paper to consider all the possible situations that could be encountered in mechanical applications. In this way, it is expected that our analysis and the prediction model developped in the next section remain valid for a wide range of volume fraction including [40\%,70\%] that corresponds to concrete.
\end{remark}





The two local indicators are therefore relevant to characterize and discriminate the local behavior of a crack. They are integrated into a probabilistic prediction model that is described in the next section.
%

\section{Markov chain model for crack prediction }
\label{new_local}
This section is devoted to the construction of the fast-to-evaluate surrogate model for crack path prediction. The crack is modelled by a Markov chain with a set of parameters.
For given crack tip and direction of propagation, several points can be reached and one of them will be selected by the model as the next crack tip. The Markov chain model proposed in this section allows to associate a probability to each of these points. The future crack tip is then chosen by a random draw based on these probabilities. The key point of this type of model is the construction of a transition kernel that precisely defines the probability for the crack tip at position $x$ to reach a new point $y$.
\\
In this section, we fully describe the development of the model with a specific transition kernel involving the two indicators presented in the section \ref{sec:indic}. The parameters are estimated in section \ref{para_estim}. Finally, section \ref{sec:pred} fully describes the crack path prediction using the proposed model. 


\subsection{Construction of the model}
\label{sec_MC}

The prediction model is based on a Markov chain model. This type of model relies on the property that the prediction of future states of a system only depends on its present state. It is therefore particularly well suited to the modeling of the crack whose propagation only depends on local information at the crack tip and not on the whole path before the tip. Let us recall the definition \ref{def_markov} of a Markov chain. For more details on Markov chains, the reader is invited to refer to the first chapter of \cite{markov_an} or any text book on Markov Chains. 
\begin{definition}[A Markov chain]:
\label{def_markov}a sequence of random variables $\left(X_n\right)_{n \in \mathbb{N}} $ with values in a set $\mathcal{E}$ is a Markov chain of state space $\mathcal{E}$ if 
for all $ k\in \mathbb{N}$, $\forall \left(x_0,...,x_{k+1}\right) \in$ $\mathcal{E}^{k+2}$ such that $P \left(X_0=x_0,\ldots,X_k=x_k \right)>0$,

\begin{eqnarray*}
P     \left(X_{k+1}=x_{k+1}|X_0=x_0,...,X_k=x_k\right) =P \left(X_{k+1}=x_{k+1}|X_k=x_k\right).
\end{eqnarray*}

Here, $P \left(X_{k+1}=y|X_k=x\right)$ denotes the conditional probability of  $X_{k+1}=y$ given $X_{k}=x$. We set $\nu_0 \left(x_0\right) =P \left(X_0=x_0 \right)$, so that  for all $ \left(x_0,...,x_n \right) \in$  $\mathcal{E}^{n+1}$, \begin{eqnarray*}
P\left(X_0=x_0,...,X_n=x_n \right) =\nu_0 \left(x_0 \right) \Pi^{n-1}_{k=0}P \left( X_{k+1}=x_{k+1}|X_k=x_k \right).\end{eqnarray*}\\
Moreover, the chain is said to be homogeneous if for all $ k\in \mathbb{N}$ and for all $\left(x,y\right) \in \mathcal{E}^2$, $P\left(X_{k+1}=y|X_k=x\right) =P\left(X_1=y|X_0=x\right)$. 

 \end{definition}

The discretization points of the crack are assumed to be a realization of homogeneous Markov chain. The discussion of Section \ref{indicators} leads to propose that the probability of transition is evaluated from a transition kernel depending on the two local geometrical indicators introduced in Definition \ref{def_descr}. 
The following modelling defines the general structure of transition probability between two points in the crack. 
\begin{Modeling}[Transition probability of crack]: for any microstructure $E$, we suppose that $\left(X^E_i\right)_{i=1,...,m_E}$ is a sequence of random variables that constitutes a homogeneous Markov chain. $(x^E_i)_{i=1,\hdots,m_E}$  defines a sequence of realizations of the Markov chain and the transition kernel of the chain defining the probability of transition from $x$ to $y$ is given by: 
\begin{eqnarray}
\label{model_init}
P_E\left(X^E_{i+1}=y|X^E_i=x\right)=c_{x,\Lambda} f_{\Lambda}\left(\tilde d_{x}\left(y\right),\tilde \theta_{x}\left(y\right)\right),
\end{eqnarray}

where $P_E$ denotes the conditional probability given $E$, $c_{x,\Lambda}$ is a normalization constant in order to define a probability and $f_{\Lambda}$ is a function parametrized by a set $\Lambda$ of parameters to be determined. 
We set the first realization $X_0=x_0$ : it is the starting point of the crack and is chosen in the microstructure.
In particular we have, 

\begin{eqnarray}
\label{MC_crack}
P_E\left(X^E_0=x_0, \dots, X^E_{m_E}=x^E_{m_E}\right)= \Pi_{i=0}^{m_E-1} P_E\left(X^E_{i+1}=x^E_{i+1}|X^E_i=x^E_i\right).
\end{eqnarray}

\end{Modeling}

The aim of this model is to reproduce the crack behaviour by weighting  the probability of propagation of the crack with the importance of the indicators according to the local configuration at the tip of the crack in the microstructure. 
Since the presence of candidate points on the aggregate sides influences the
local direction of propagation, the expression of $f_{\Lambda}$ depends on the configuration (F1 or F2). We use a decreasing exponential function in order to penalize high values of the indicators.



Based on the modelling \ref{model_init} and our assumptions, the following model presents the final transition probability by specifying $c_{x,\Lambda}$ and $f_\Lambda$.

\begin{Modeling}[Probabilistic cracking model]:
\label{MC1}let $x$ be a discretization point reached by the crack and belonging to an aggregate A. If $\{y^E_k\}_{k=1,\dots,K_E}$ is the set of candidate points in  the field of view of  $x$, then the transition kernel is given by:
\vskip.1in
\begin{itemize}

\item Configuration F1 (\figurename~\ref{fig:F1}): let
$\Lambda_{F1}=\left(\mu_1,\mu_2,\mu_3,\mu_4,\mu_5,\mu_6\right)$ be the  parameters associated to this configuration. If $\{y^E_k\}_{k=1,\dots,r_E}$ defines the set of candidate points on $A$ and $\{y^E_k\}_{k=r_{E}+1,\dots,K_E}$ the candidate points on the other aggregates, $c_{x,\Lambda}=c_{x,\Lambda_{F1}}$ and $f_{\Lambda}=f_{\Lambda_{F1}}(\tilde{d}_x(y),\tilde{\theta}_x(y))$ with :
\begingroup
\large
\begin{eqnarray}
f_{\Lambda_{F1}}(\tilde{d}_x(y),\tilde{\theta}_x(y))= \left\{
    \begin{array}{ll}
        e^{-\mu_1 (\tilde \theta_{x}(y))^{\mu_2}} & \mbox{if } y \in A \\
         e^{-\mu_3 (\tilde d_{x}(y) \tilde \theta_{x}(y))^{\mu_6}-\mu_4 (\tilde d_{x}(y))^{\mu_5}} & \mbox{else}
    \end{array}
\right.
\end{eqnarray}
\endgroup
and the normalizing constant $c_{x,\Lambda_{F1}}$ is thus,
\begingroup
\large
\begin{eqnarray}
c_{x,\Lambda_{F1}}=\frac{1}{\sum_{k=1}^{K_E}f_{\Lambda_{F1}}(\tilde{d}_x(y),\tilde{\theta}_x(y))}
\end{eqnarray}
\endgroup

\item Configuration F2 (\figurename~\ref{fig:F2}): let $\Lambda_{F2}=\left(\lambda_1,\lambda_2,\lambda_3,\lambda_4,\lambda_5,\lambda_6\right)$ be the parameters associated to this configuration, then $c_{x,\Lambda}=c_{x,\Lambda_{F2}}$ and $f_{\Lambda}=f_{\Lambda_{F2}(\tilde{d}_x(y),\tilde{\theta}_x(y))}$ with :

\begingroup
\large
\begin{eqnarray}
f_{\Lambda_{F2}}(\tilde{d}_x(y),\tilde{\theta}_x(y))=e^{-\lambda_1 (\tilde d_{x}(y) \tilde \theta_{x}(y))^{\lambda_5}-\lambda_2 (\tilde d_{x}(y))^{\lambda_6}-\lambda_3(\tilde \theta_{x}(y))^{\lambda_4}}
\end{eqnarray}
\endgroup
and the normalizing constant $c_{x,\Lambda_{F2}}$ is thus,
\begingroup
\large
\begin{eqnarray}
c_{x,\Lambda_{F2}}=\frac{1}{\sum_{k=1}^{K_E}f_{\Lambda_{F2}}(\tilde{d}_x(y),\tilde{\theta}_x(y))}
\end{eqnarray}
\endgroup

\end{itemize}

\end{Modeling} 

The different variables integrated in this modeling come from the analysis performed in Section \ref{sec_app_ind}. 
The mixed term allows to take values of distance and angle of the same order of magnitude.
The distance term allows to take into account the neighboring attraction of the aggregates. 
The angle term allows to take into account the direction of propagation.
In the configuration $F_1$, the distinction between the possibility of following the aggregate and crossing the matrix is taken into account in the expression of  $f_{\Lambda_{F_1}}$ that depends on the two previous situations. Moreover, as observed in section \ref{sec_app_ind} for candidate points on the same aggregate, the distance has no influence and therefore, only the angle indicator is included in the model. For candidate points that require crossing the matrix in the configuration F1, experimentally, it is the presence of an aggregate close to the crak tip that has the most important influence on the change of crack behavior. That is why the angle indicator is only taken into account in the interaction term contrarly to the transition kernel in configuration F2. It allows reducing the number of parameters in the model and the variability of their estimate. \\

More generally, in both configurations F1 and F2, the interaction term between distance and angle allows integrating the effect of the combination of small values of the two indicators that was observed in section \ref{sec_app_ind}. It constrains the crack to follow a propagation direction while keeping the attraction of the closest aggregates.\\

The model parameters are estimated from a training set of microstructures whose cracking has been simulated with XPER code. Under the independence assumption of 
the microstructures, the estimate is performed by maximization of the likelihood. More precisely, if \noindent $\{E_i\}_{i=1,\ldots,Q}$ denotes the training set of microstructures and  $(x^{E_i}_j)_{j \in I_{E_i}^{F1}}$ et $(x^{E_i}_j)_{j\in I_{E_i}^{F2}}$ are the sequences of $I$ points selected by the crack on the microstructure $E_i$ in both configurations, the parameters $\Lambda_{F1}^\star$ and $\Lambda_{F2}^\star$ are solutions of: 
\begin{itemize}
\label{max_conf}
\item Configuration F1:
   \begin{eqnarray}
\label{ML_form}
\Lambda^{\star}_{F1}=\argmax_{\left(\mu_1,\ldots,\mu_6\right)} \left\{ \Pi_{i=1}^Q \Pi_{j \in I_{E_i}^{F1} } P_{E_i}\left(X_{j+1}^{E_i}=x_{j+1}^{E_i} | X_{j}^{E_i}=x_{j}^{E_i}\right) \right\}
\end{eqnarray}

\item Configuration F2:
   \begin{eqnarray}
\label{ML_form1}
\Lambda^{\star}_{F2}=\argmax_{\left(\lambda_1,\ldots,\lambda_6\right)} \left\{ \Pi_{i=1}^Q \Pi_{j \in I_{E_i}^{F2} } P_{E_i}\left(X_{j+1}^{E_i}=x_{j+1}^{E_i} | X_{j}^{E_i}=x_{j}^{E_i}\right) \right\}
\end{eqnarray}

\end{itemize}

\noindent where $P_{E_i}$ is given by Modeling (\ref{MC1}). 



\subsection{Cracking prediction with the probabilistic model}
\label{sec:pred}
 
This section describes the crack path prediction for a given microstructure using the Markov chain model introduced in Modeling \ref{MC1}. The parameters of the transition kernel are assumed to be known, their estimate is studied in Section \ref{sec_appl}. This prediction relies on a procedure that starting from the crack tip, provides the next point of the field of view that is reached by the crack.

More precisely, the local indicators introduced in Definition \ref{def_descr} are first evaluated integrating Procedure \ref{prop_penombre} to reduce the set of candidate points in the field of view. Then, the Markov chain model can evaluate the probability of each point of the set to be 
the next point reached by the crack. Finally, a realization is retained according to the evaluated probability to select the next point of the crack. This procedure is described below:

\begin{procedure}[Process of a local prediction to determine the next point of the crack]:
\label{prop_pred0}given $x^E_i$ the crack tip in a microstructure $E$ and $\overrightarrow{u_{x^E_i}}$ the local propagation direction, the prediction includes three steps:

\begin{itemize}

\item Step 1: construct the field of view of $x^E_i$, identify the candidate points following Procedure \ref{prop_penombre} ($\{y^E_k\}_{k \in 1,\ldots,K_E}$ is the set of remaining points) and evaluate the normalized indicators $\tilde d_{x^E_i}\left(y^E_k\right)$ and $\tilde \theta_{x^E_i}\left(y^E_k\right)$ $\forall k$.

\item Step 2: according to the configuration, compute $\{P_{E}\left(X^E_{i+1}=y^E_k| X^E_i=x^E_i\right)\}_{k=1,\ldots,K_E}$ using Modeling \ref{MC1}.

\item Step 3: select a realization from the discrete law $\Sigma_{k =1}^{K_E} P_{E} \left(X^E_{i+1}=y^E_k| X^E_i=x^E_i\right) \delta_{y^E_k}$.

\end{itemize}

\end{procedure}  

In this procedure, the objective is to randomly drawn the next point of the crack among the points with the highest probabilities and not to select the candidate point with the highest probability.  It is considered that several points can be good candidates for to be the next point of crack.
Starting from the  crack tip (initial position), this procedure is successively applied to any new point on the crack until the boundary of the domain is reached.

Thus the prediction model is stochastic. Starting from the same initial position, it can be applied to obtain several realizations of the crack path. Therefore, it allows to quantify the uncertainty associated to the prediction and to the mechanical quantities of interest.
In the next section~\ref{sec_appl}, tools to use this set of realizations to determine the most optimal crack path will be presented. 

\begin{algorithm}[H]
  \caption{Local prediction}\label{ALG}
  \begin{algorithmic}[1]
    \State E is the discrete granular microstructure
    \State $x^E$ is the point of crack tip
    \State DirectionProp is the local direction in mode I
    \State MarkovChainModel is the model in \ref{MC1} (gives the probabilities for the candidate points)
    \State RandWeighted performs a weighted draw for select the next point of the crack 
    \State ParamModel are the estimated parameters of the prediction model
    \Function{Cracking}{$E, x_0^E, ParamModel,DirectionProp$}
     \State FieldviewAgg = set of aggregates in the field of view of the
point $x^E$ associated to DirectionProp
    \State FieldviewPoint = set of points located on the aggregates of
FieldviewGran 
       \For {G in FieldviewAgg } 
            \State ShadowG= {shadow zone of aggregate G}
            \For {point in FieldviewPoint  } 
                \If {point in ShadowG }
                    \State Delets the point in FieldviewPoint
                \EndIf
            \EndFor
        \EndFor

    \State {ProbaPoint=MarkovChainModel(FieldviewPoint,ParamModel)}
    \State {PositionPointChoosen=RandWeighted(ProbaPoint)}
    \\
    \hspace{0.5cm}  \Return  PositionPointChoosen
    \EndFunction
  \end{algorithmic}
\end{algorithm}

\begin{figure}[h!]
    \centering
    \includegraphics[scale=0.65]{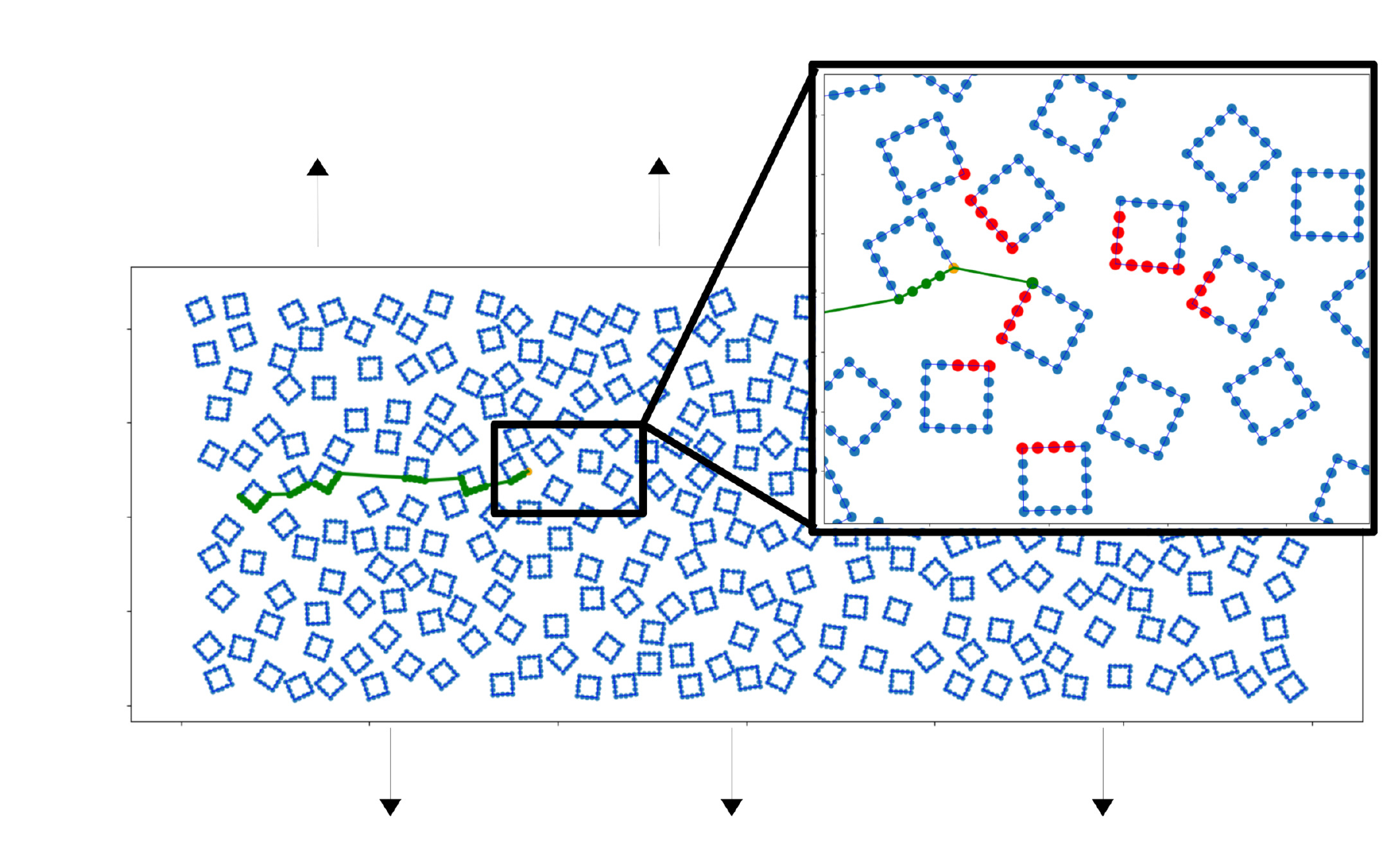}
    \caption{Illustration of the prediction (procedure \ref{prop_penombre}) in the F2 configuration with a zoom on the region of interest. The field of view is orientated with respect to a propagation direction orthogonal to the stress effort. The orange point is the crack tip, the red points are candidate points in the field of view where the probability to be reached is evaluated, the green points are the candidate points selected by the model}
    \label{fig:foc_pred}
\end{figure}

\figurename~\ref{fig:foc_pred} provides an illustration of the prediction procedure in the F2 configuration. The algorithm \ref{ALG} summarizes the three steps of the procedure.

\newpage
\section{Numerical application}
\label{sec_appl}
Several types of numerical applications are considered in this section. They first concern the estimate of the parameters (Section \ref{para_estim}) of the transition kernel. Then, we focus on the ability of the model to correctly predict the crack path (Section \ref{crack_pred}) and the associated mechanical quantities of interest. 

\subsection{Parameter estimate }
\label{para_estim}
 The parameter estimate exploits a training set of 35 microstructures  numerically cracked with the XPER code from the test case described in Section \ref{lim_mech} with square shape aggregates. The estimation of the parameters of the Markov chain are obtained by maximising the likelihood (Equations (\ref{ML_form}) and (\ref{ML_form1})). The maximisation is performed with an optimized version of the BFGS algorithm (\cite{LBFGS}).  \\
 
 \tablename~\ref{tab_estimate} provides the estimated values for $\Lambda_{F1}$ and $\Lambda_{F2}$. 

 \begin{table}[!h]

\begin{center}
\begin{tabular}{|c|c|}
  \hline
  & Estimated values  \\
  \hline
 $\mu_1$ &  7.06\\
 $\mu_2$  & 4.1 \\
  $\mu_3$ &  30.2\\ 
   $\mu_4$ &  8.9\\ 
    $\mu_5$ &  0.2\\ 
     $\mu_6$ &  0.85 \\ 
  \hline
  \hline
 $\lambda_1$ &  34.2\\
 $\lambda_2$  & 9.2 \\
  $\lambda_3$ &  13.16\\ 
   $\lambda_4$ &  1.79\\ 
    $\lambda_5$ &  1.08\\ 
     $\lambda_6$ &  0.42 \\ 
  \hline
\end{tabular}
\caption{Estimated parameters for $\Lambda_{F1}$ and $\Lambda_{F2}$, $\Lambda_{F1}=\left(\mu_1,\hdots,\mu_6\right)$ and $\Lambda_{F2}=\left(\lambda_1,\hdots,\lambda_6\right)$ }

\label{tab_estimate}
\end{center}

\end{table} 
 
Let us recall that in Modeling \ref{MC1}, the parameters $\mu_1$, $\mu_3$, $\mu_4$, $\lambda_1$, $\lambda_2$, $\lambda_3$ correspond to multiplicative factors, the others are exponents.

In the configuration F1, the two highest values of the multiplicative factors ($\mu_3$ and $\mu_4$) and the two lowest values of the exponents ($\mu_6$ and $\mu_5$) correspond respectively to the parameters of the distance and the interaction terms. These two quantities are associated to candidate point located on another aggregate. As a result, the crack will tend to favour a path on the same aggregate unless the angle to stay on the aggregate is high and the crossing of the matrix is associated to low distances and angles. This is consistent with the analysis of Section \ref{sec_app_ind}. \\

\begin{figure}[h!]
\centering
    \includegraphics[scale=0.7]{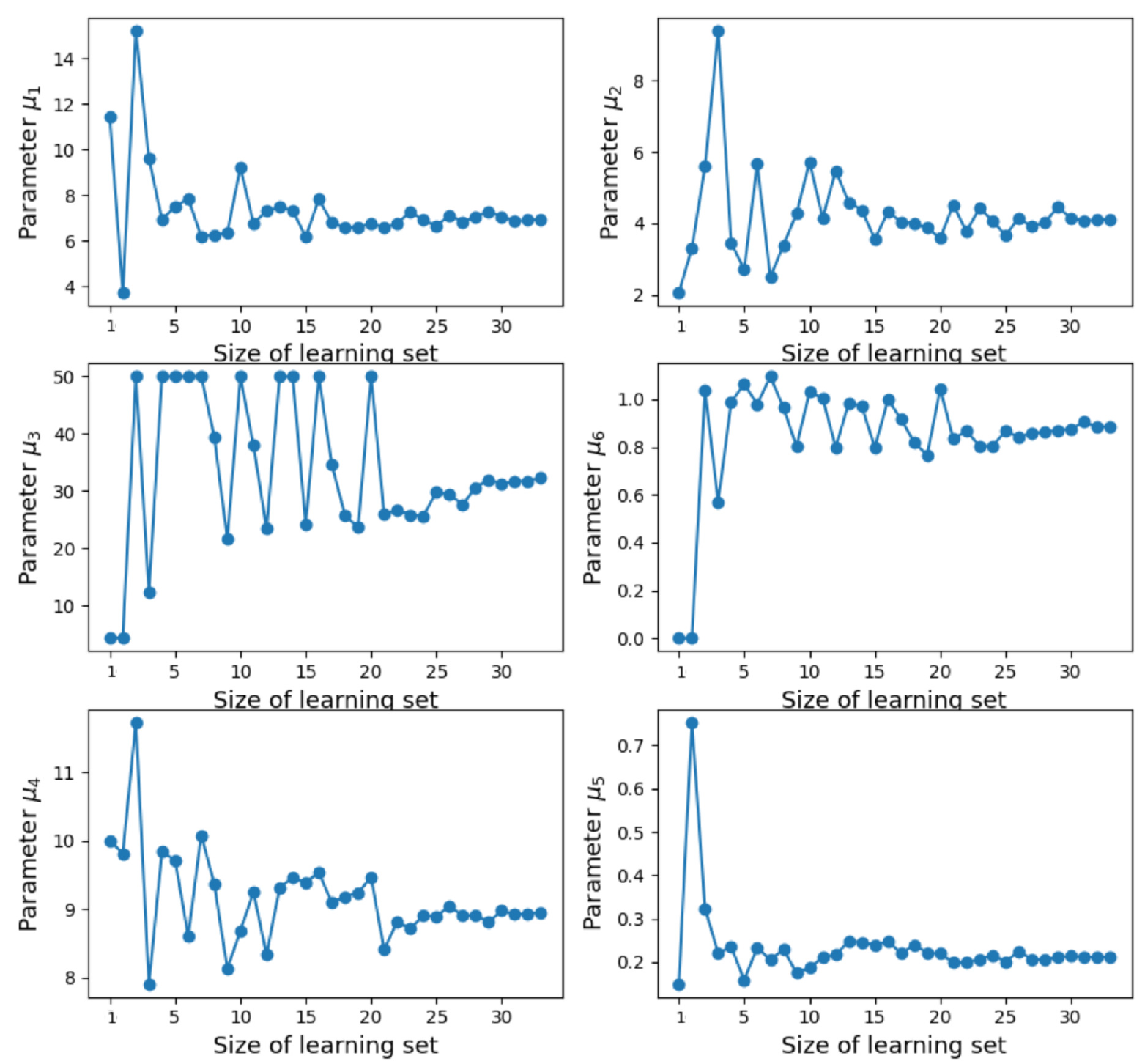}
    \caption{Parameter estimate with respect to the size of the training set in the F1 configuration. From top to bottom, parameters associated to $\tilde \theta_x$ (points on the same aggregate), to the interaction term (points requiring crossing the matrix) and to $\tilde d_x$ (points requiring crossing the matrix)}
    \label{fig :var_para_mat}
\end{figure}

\begin{figure}[h!]
\centering
    \includegraphics[scale=0.7]{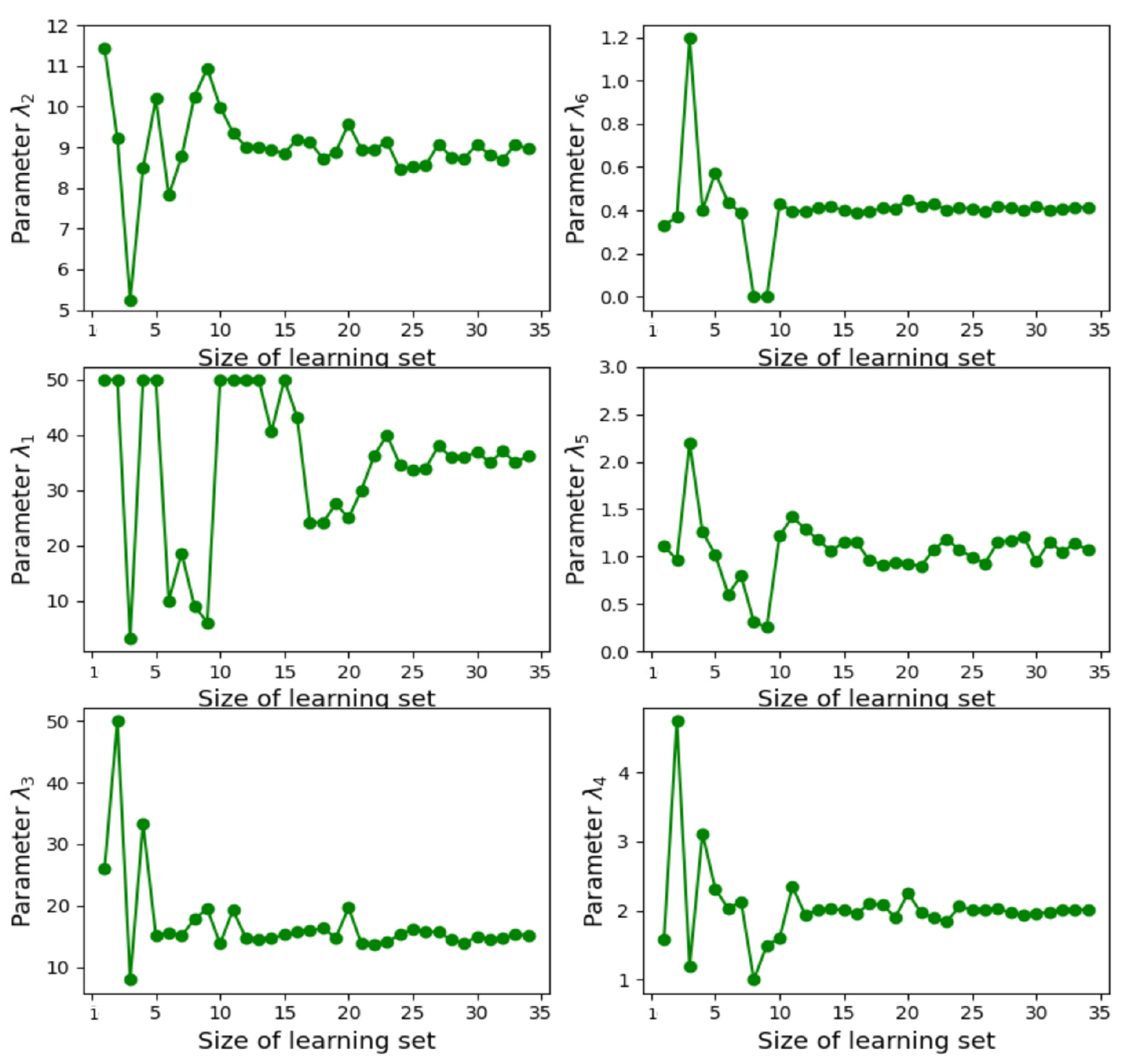}
    \caption{Parameter estimate with respect to the size of the training set in the F2 configuration. From top to bottom, parameters associated to $\tilde d_x$, to the interaction term and to $\tilde \theta_x$ }
    \label{fig :var_para_granmix}
\end{figure}

It is also interesting to study the variability of the estimate with respect to the size of the training set. More precisely, the parameter estimate is performed from a training set including an increasing number of microstructures.
\figurename~\ref{fig :var_para_mat} and \figurename~\ref{fig :var_para_granmix} display the estimate of each parameter. It is possible to observe a first stabilization for the majority of the parameters from about ten microstructures. In all cases, beyond 25 microstructures, the estimate can be considered to be stabilized for all parameters.  \\


\subsection{Cracking prediction}
\label{crack_pred}
We evaluate the performances of the prediction model with the estimated parameters provided by  \tablename~\ref{tab_estimate}. Since the model is stochastic, for a given microstructure, its output is a random crack path. When run $M$ times, we obtain a set of $M$ cracks.  \\

In this section, the model is applied to the training set for verification and to a test set (i.e. not used for the parameter estimate) of $30$ microstructures with multiform aggregates (aggregates represented by regular polygons with a number of sides varying from 3 to 8) for validation.  It is important to keep in mind that the change of aggregate shapes between the training and the test sets does not require new XPER simulations to construct the prediction model, and therefore there is no extra significant computational time for the crack prediction. \\

We first focus on the analysis of the crack paths then consider the prediction of the tortuosity.

\subsubsection{Crack path }
For a given microstructure $E$, we denote by $\mathcal{X}_E^k=( x^E_{i,k})_{i=1,\dots,m_E^k}$ the sequence associated to the $k^{th}$ ($k=1,\dots,M$) crack realization. \\

\paragraph{ Median path }
When the objective is to derive a unique crack path that approximates the deterministic computer code simulation at a low cost,  we use a median path from the set of points $\mathcal{X}_E^\star$ satisfying:

\begin{eqnarray}
\mathcal{X}_E^\star=\argmin_{\{\mathcal{X}_E^k\}_{k=1,\ldots,M}} \left(\Sigma_{j=1,j\ne k}^{M} D\left(\mathcal{X}_E^k,\mathcal{X}_E^j\right)\right)
\end{eqnarray}

where $D$ is the Fr\'echet distance (\cite{frechet,measure}) which is defined as follows :
\begin{definition}[Fréchet distance]: let $X_1$ and $X_2$ be two crack paths.
The Fréchet distance between two crack is defined as follows :

\[
D(X_1,X_2)=\inf_{\alpha,\beta} \max_{t\in[0,1]} || X_1(\alpha(t)) -X_2(\beta(t))||
\]

where $X_1, X_2 : [0,1] \rightarrow \mathbb{R}^2$ are parametrizations of the two crack and $\alpha,\beta :[0,1] \rightarrow [0,1]$
\end{definition}
The median crack is considered the most optimal crack for a given microstructure.

\paragraph{ Confidence region }
To quantify the uncertainty of the median, we define a confidence region.
The region is constructed from point-value evaluations of percentiles. This type of construction requires a parametrization of each predicted path (see definition \ref{def_fiss}). In this test case, the crack is initialized on the left hand side boundary of the microstructure and propagates until the failure of the microstructure.

For any $x^1 \in [0,L]$, the uncertainty is quantified by estimating the $5\%$ and $95\%$ percentiles from the sample $\left(h_E^1\left(x^1\right),\ldots,h_E^{M}\left(x^1\right)\right)$. Denoting $\left(\tilde h_E^1\left(x^1\right),\ldots,\tilde h_E^{M}\left(x^1\right)\right)$ the ordered sample (increasing order), the uncertainty interval at $x^1$ is defined by  $\left[\tilde h_E^{\lfloor 0.05M \rfloor}\left(x^1\right),\tilde h_E^{\lceil 0.95M \rceil}\left(x^1\right)\right]$ (where $\lceil . \rceil$ is the ceiling and $\lfloor .  \rfloor$ the floor) which constitutes the upper curve and the lower curve. 
 The region of confidence is finally taken as the convex hull and the uncertainty of the model prediction can be quantified by the diameter of this region computed as the Fr\'echet distance between the lower and upper curves defining this hull. 
The confidence region gives the area where the crack has the highest probability of passing. 
    \\

\begin{figure}[h]
    \centering
    \includegraphics[width=\textwidth]{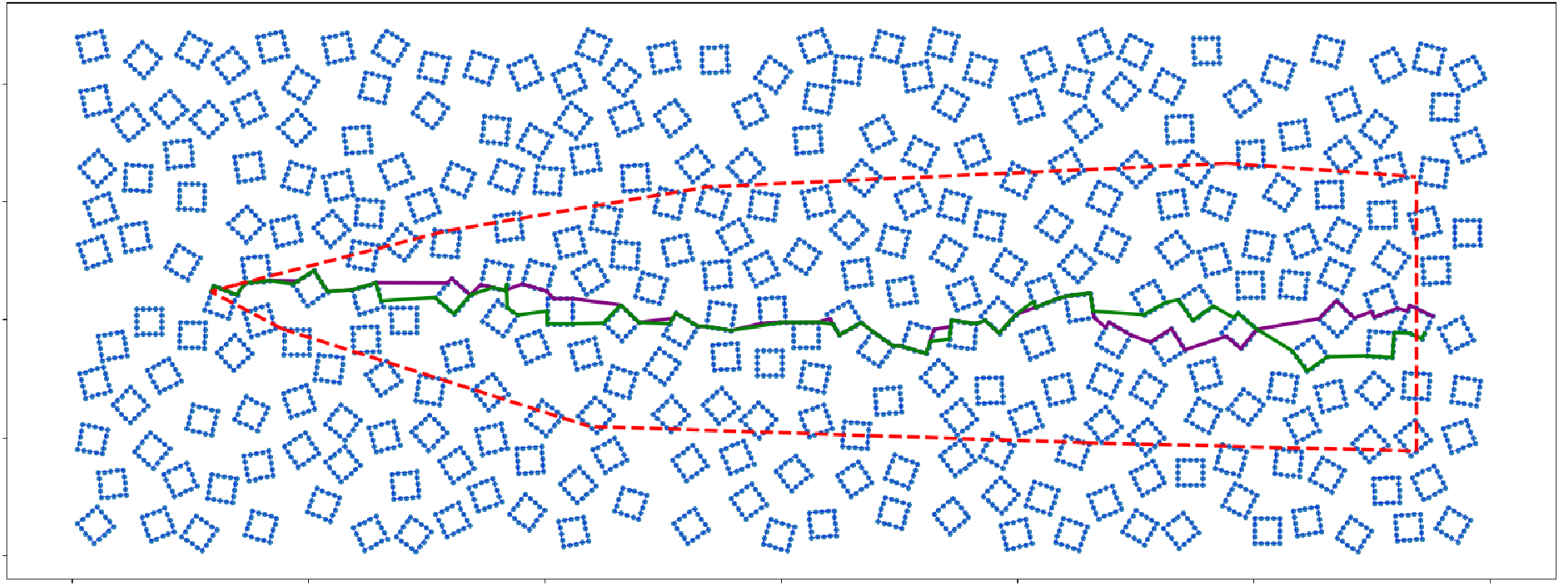}
    \caption{Example of path prediction for square aggregates: XPER simulation (purple), median path (green), region of confidence (dotted red curve)}
    \label{fig:conv_crack}
\end{figure}

\begin{figure}[h]
    \centering
    \includegraphics[width=\textwidth]{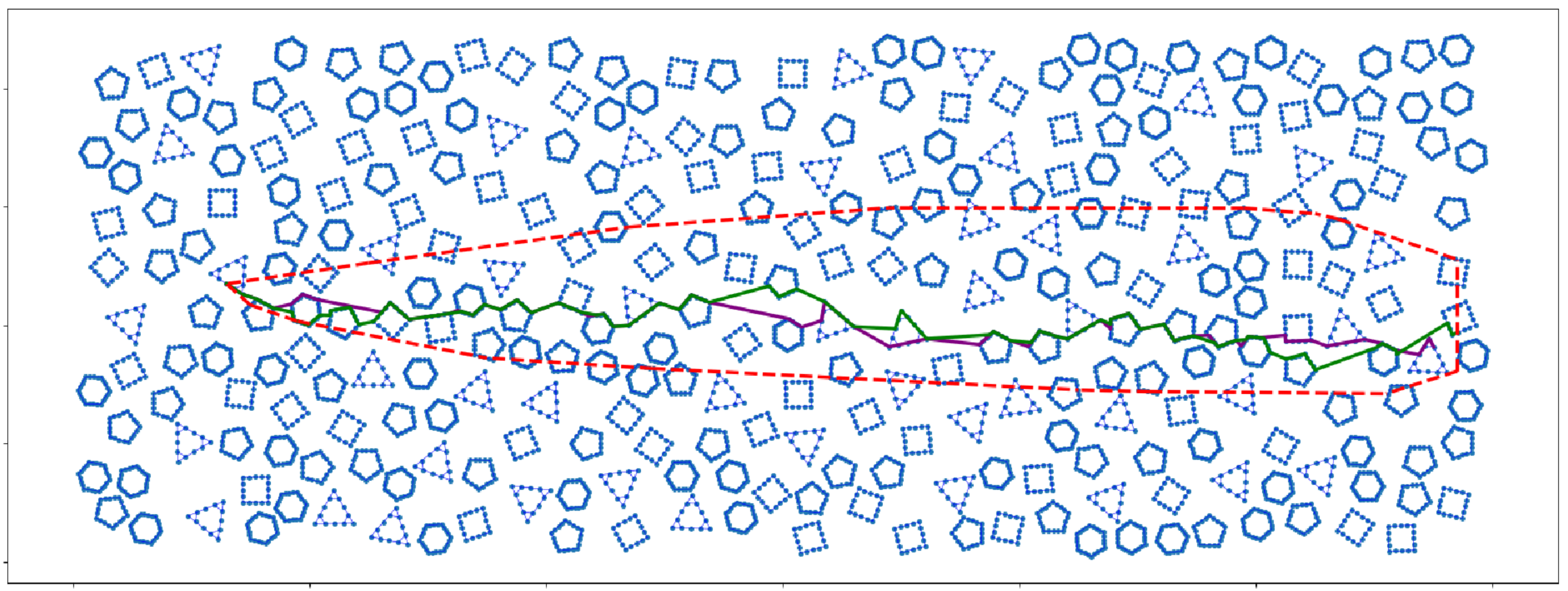}
    \caption{Example of path prediction for multiform aggregates: XPER simulation (purple), median path (green), region of confidence (dotted red curve) }
    \label{fig:conv_crack2}
\end{figure}

\figurename~\ref{fig:conv_crack} and \figurename~\ref{fig:conv_crack2} provide two examples of median crack path and confidence region as well as a comparison with the XPER simulation for a microstructure of the training set (only square inclusions) and a microstructure of the test set (various shape inclusions).

\begin{figure}[h!]
    \centering
    \includegraphics[width=\textwidth]{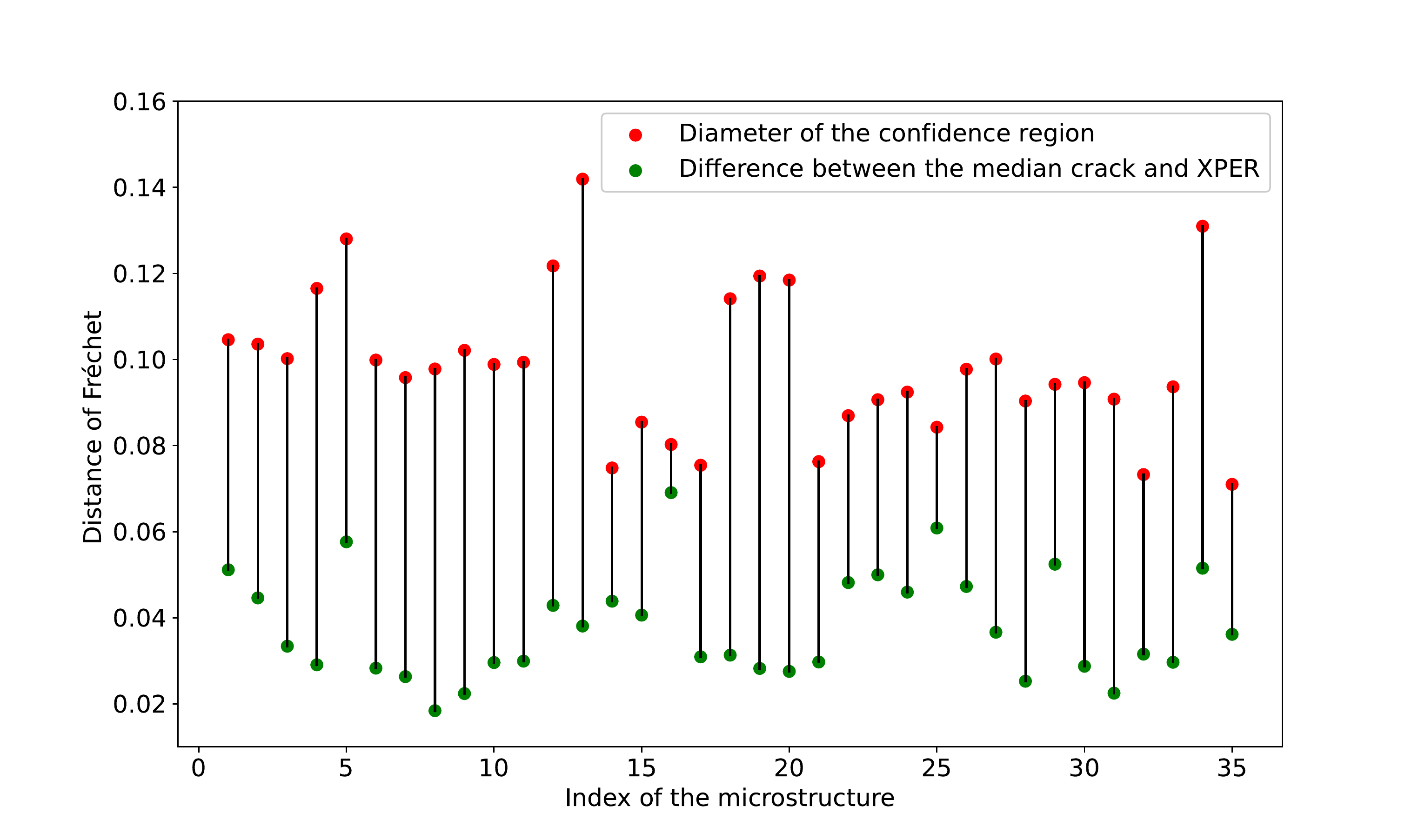}
    \caption{Prediction on the training set (square aggregates): Fr\'echet-distance-based error (green) between the median crack path and XPER simulation, radius of the confidence region (red)  }
    \label{fig:comp_carre}
\end{figure}

\begin{figure}[h!]
    \centering
    \includegraphics[width=\textwidth]{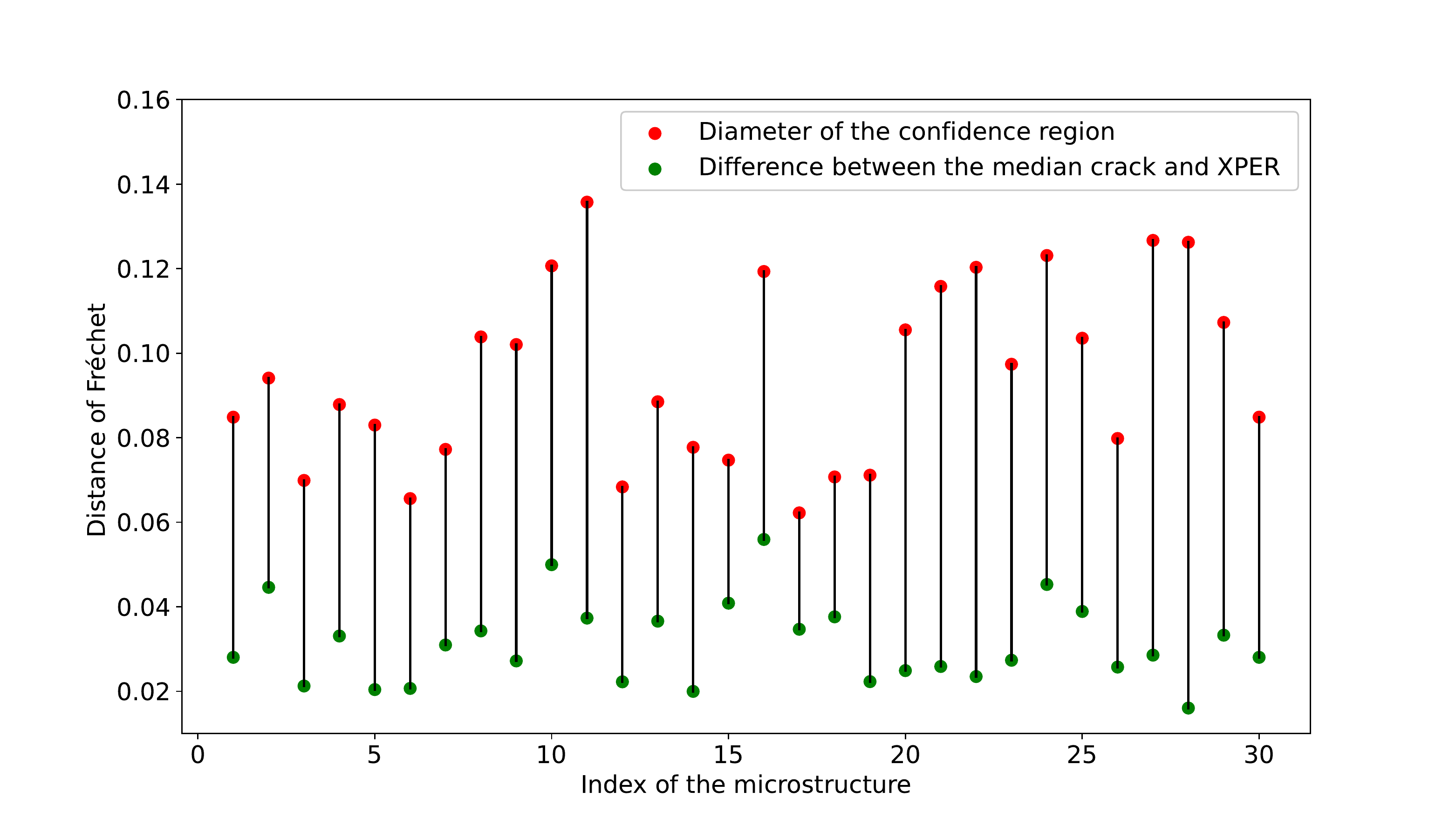}
    \caption{Prediction on the test set (multiform aggregates): Fr\'echet-distance-based error (green) between the median crack and XPER simulations, radius of the confidence region (red) }
    \label{fig:comp_mult}
\end{figure}

The number of paths calculated by the prediction model is $M=100$. To complement the previous comparison, \figurename~\ref{fig:comp_carre} and \figurename~\ref{fig:comp_mult} display the Fr\'echet-distance-based error between the XPER simulation and the median path  as well as the uncertainty for all the microstructures of the training and test sets. For a better understanding of their values, the results of \figurename~\ref{fig:conv_crack} and \figurename~\ref{fig:conv_crack2} correspond to the microstructures $31$ and $28$ on these figures. \\

\begin{figure}[h]
    \centering
    \includegraphics[width=\textwidth]{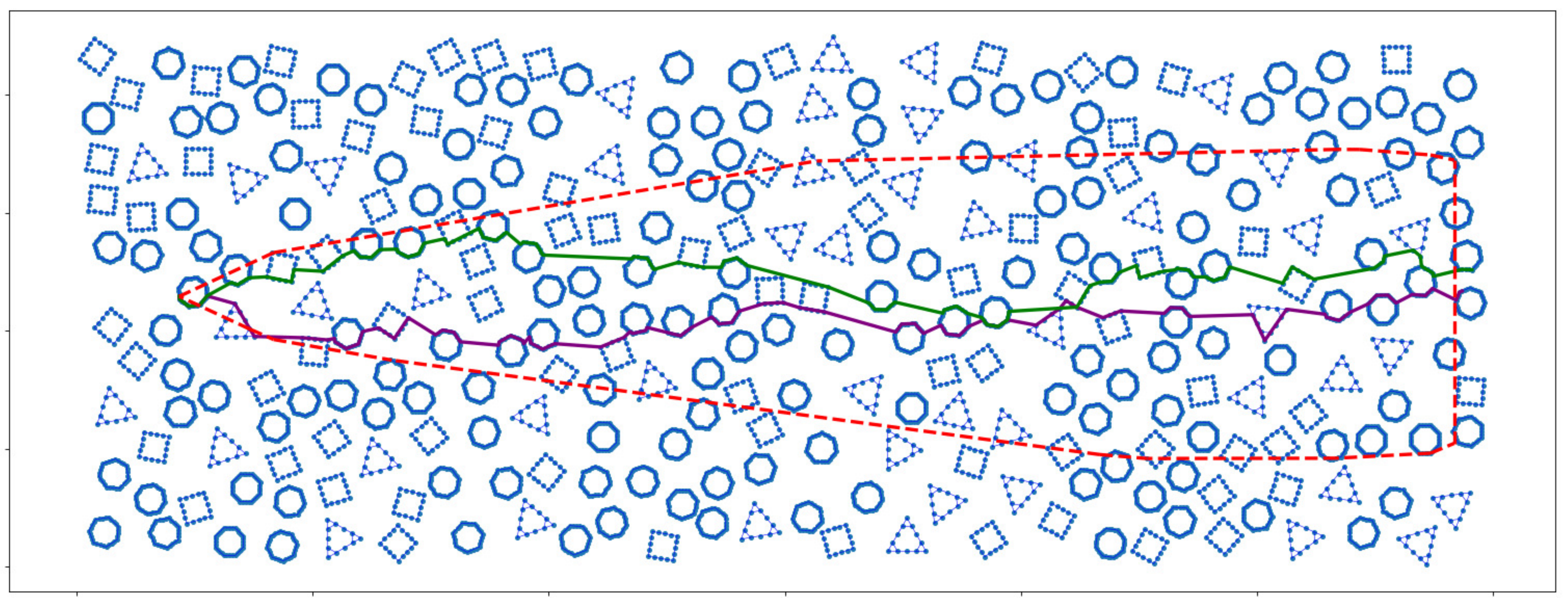}
    \caption{Example of path prediction for multiform aggregates: XPER simulation (purple), median path (green), region of confidence (dotted red curve) }
    \label{fig:bad_crack}
\end{figure}

\newpage
It can be observed that the error between the median crack path and the XPER simulation is always smaller than the uncertainty. 
The accuracy of the median prediction can be reduced for some microstructures of the training and test sets. As an example, \figurename~\ref{fig:bad_crack} shows the case associated to the largest error on the test set. 

This lack of accuracy can be explained by a situation encountered by the crack where two candidate points of the field of view have similar values of their indicators. As a result, they exhibit the same probability to be the next point on the crack path. Since our model is stochastic, these two points are therefore reached by several crack realizations and it is not surprising that, for some microstructures, the median crack path does not correspond to the same local choice than the XPER simulation. However, despite this local bifurcation, the rest of the crack path is in good agreement with the mode I propagation direction. Moreover, the XPER simulation lies in the region of confidence. Note that this result holds for both microstructures of the training and test sets.

\begin{figure}[h!]
\centering
\includegraphics [scale=0.35]{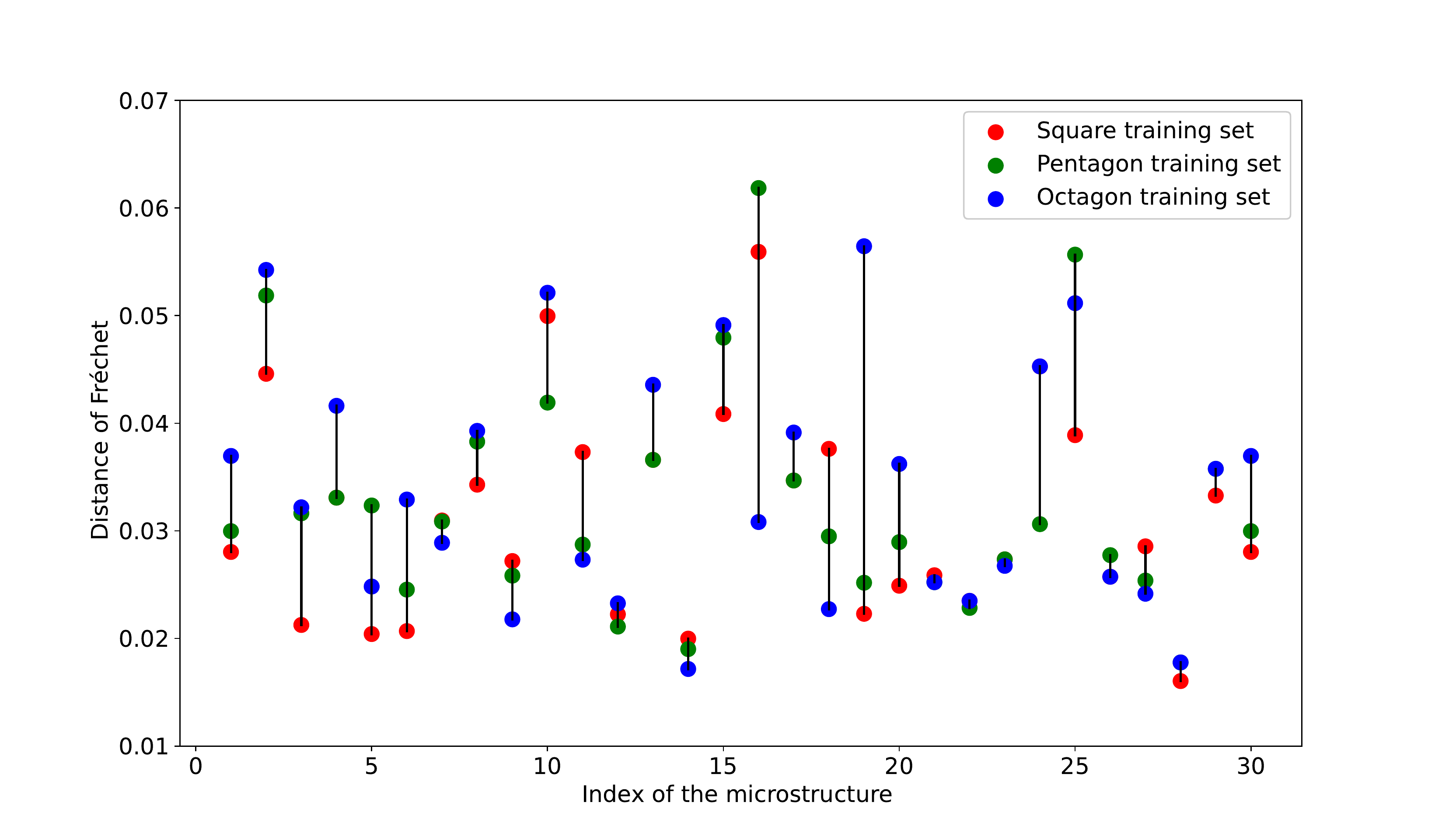}
 \caption{ Comparison of Fr\'echet distances between the median crack path and the XPER simulations for different training sets}
 \label{fig:dist_frech_BA}
\end{figure}

\paragraph{ Influence of the training set}
\label{sec:influence}
We also study the influence of training set on the predictions for microstructures of the test set. Besides the training set with square aggregates, we consider two new training sets of $35$ microstructures with pentagonal, resp. octagonal, aggregates numerically cracked with XPER code under the same conditions as those of Section \ref{lim_mech}. The parameters of the prediction model are estimated with the BFGS algorithm.
\figurename~\ref{fig:dist_frech_BA} gives the results associated to the construction of the median crack path for the 3 training sets. The set with square aggregates leads to the smallest error for the largest number of microstructures. 
However on the 30 microstructures studied, the 3 training bases show good performances on the test set with the same error magnitude.
We can therefore conclude that there is no significant influence of the shape of the training set.

\paragraph{ Evaluation of the computational time }
We conclude this section by focussing in \tablename~\ref{tab_tps_tract_simp} on the CPU time associated to the crack prediction by our model for both training and test sets.

\begin{table}[h!]
\begin{center}
\begin{tabular}{|c|c|}
\hline
One iteration of Algorithm \ref{ALG} & $0.13$ \\
\hline
One crack realization (M=1) & $13$ \\
\hline
Median crack (M=100) & $1375$ \\
\hline
\end{tabular}
\caption{Average computational time in seconds associated to the prediction using the probabilistic model in the case of a single processor}
\label{tab_tps_tract_simp}
\end{center}
\end{table}

By comparing to the XPER simulation (\tablename~\ref{tab1-3}), our model allows a drastic reduction of the computational cost (reduction by a factor $2575$ for the median crack path when considering the same number of processors).

\subsubsection{Analysis of a mechanical quantity: the tortuosity}

\begin{figure}[!h]
\center
\includegraphics[scale=0.35]{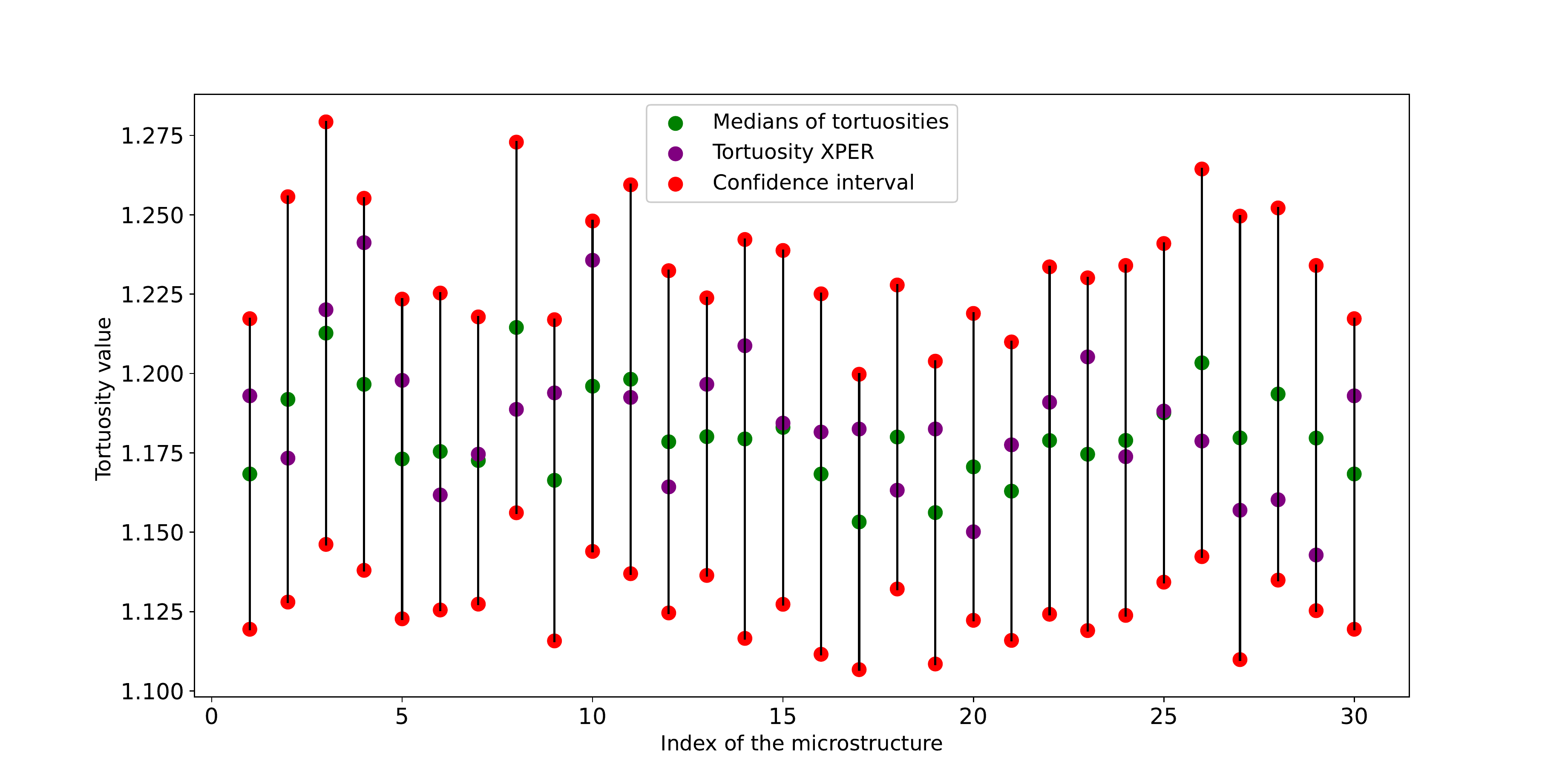}
\caption{\label{fig:inc_tort} Tortuosity prediction on the test set (multiform aggregates): XPER simulations (purple), median predictions (green), uncertainty intervals (red)}
\end{figure}

For each microstructure, the tortuosities of the $M$ predicted paths are computed and we focus on their median and uncertainty interval. This interval is obtained from percentile estimate similarly to the previous section. \figurename~\ref{fig:inc_tort} displays the results when $M=100$ for each microstructure of the test set and a comparison with the tortuosity coming from the XPER simulation is performed as well.\\  

\begin{figure}[!h]
    \centering
        \includegraphics[scale=0.45]{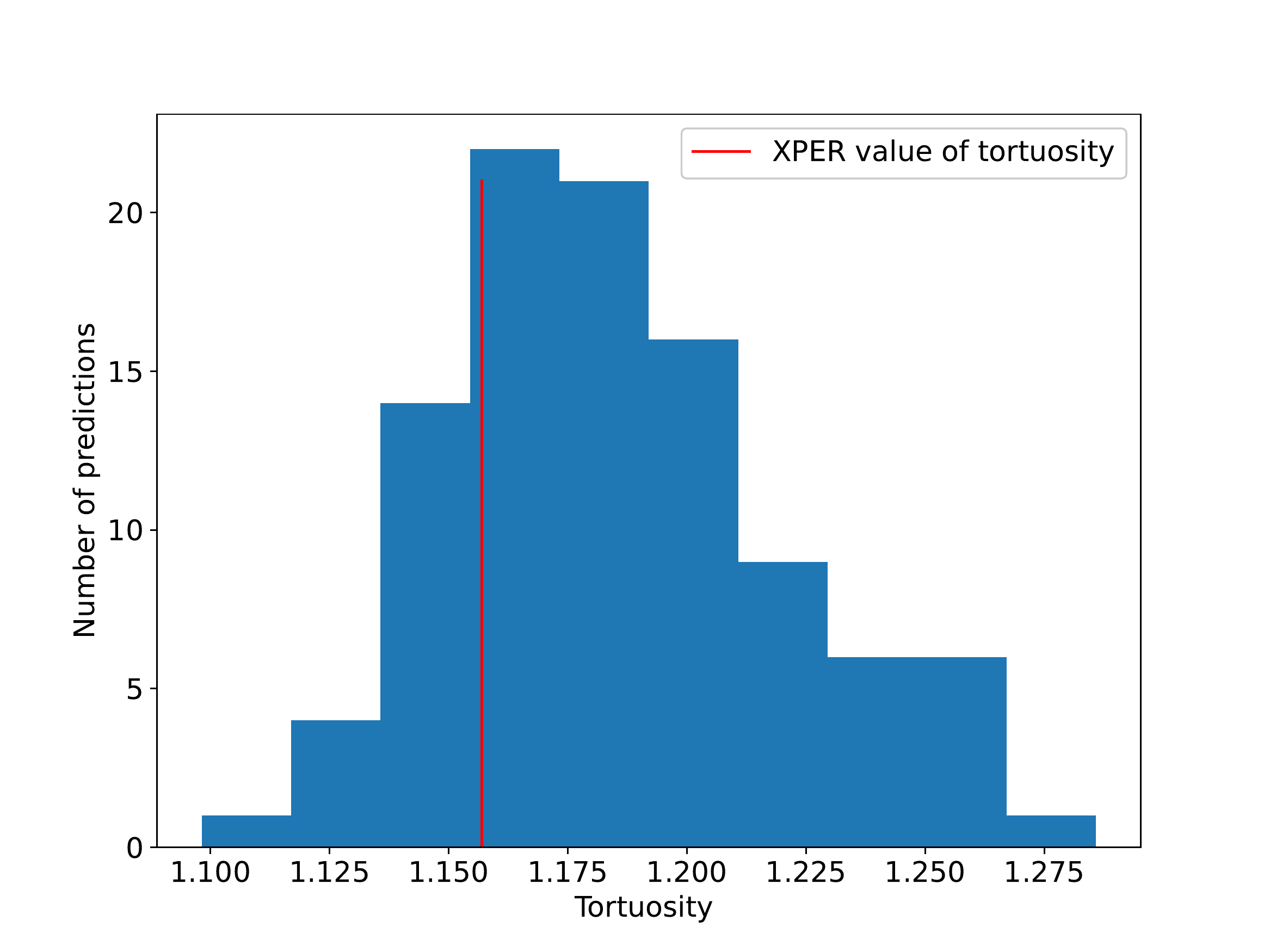}
    \caption{\label{fig:hist_faible} Tortuosity histogram for a microstructure of the test set}
\end{figure}

It is also important to keep in mind that the model provides more information than a median and a confidence interval. It allows to derive the tortuosity histogram (see for example \figurename~\ref{fig:hist_faible}) of each microstructure.

This statistical quantity can be exploited to evaluate the probability for the tortuosity to belong to a given variation range. This type of information is relevant for computational cost reduction since it allows performing targeted XPER simulations i.e. simulations leading to a tortuosity in a variation range of interest. \\

\begin{figure}[!h]
 \centering
\includegraphics[scale=0.45]{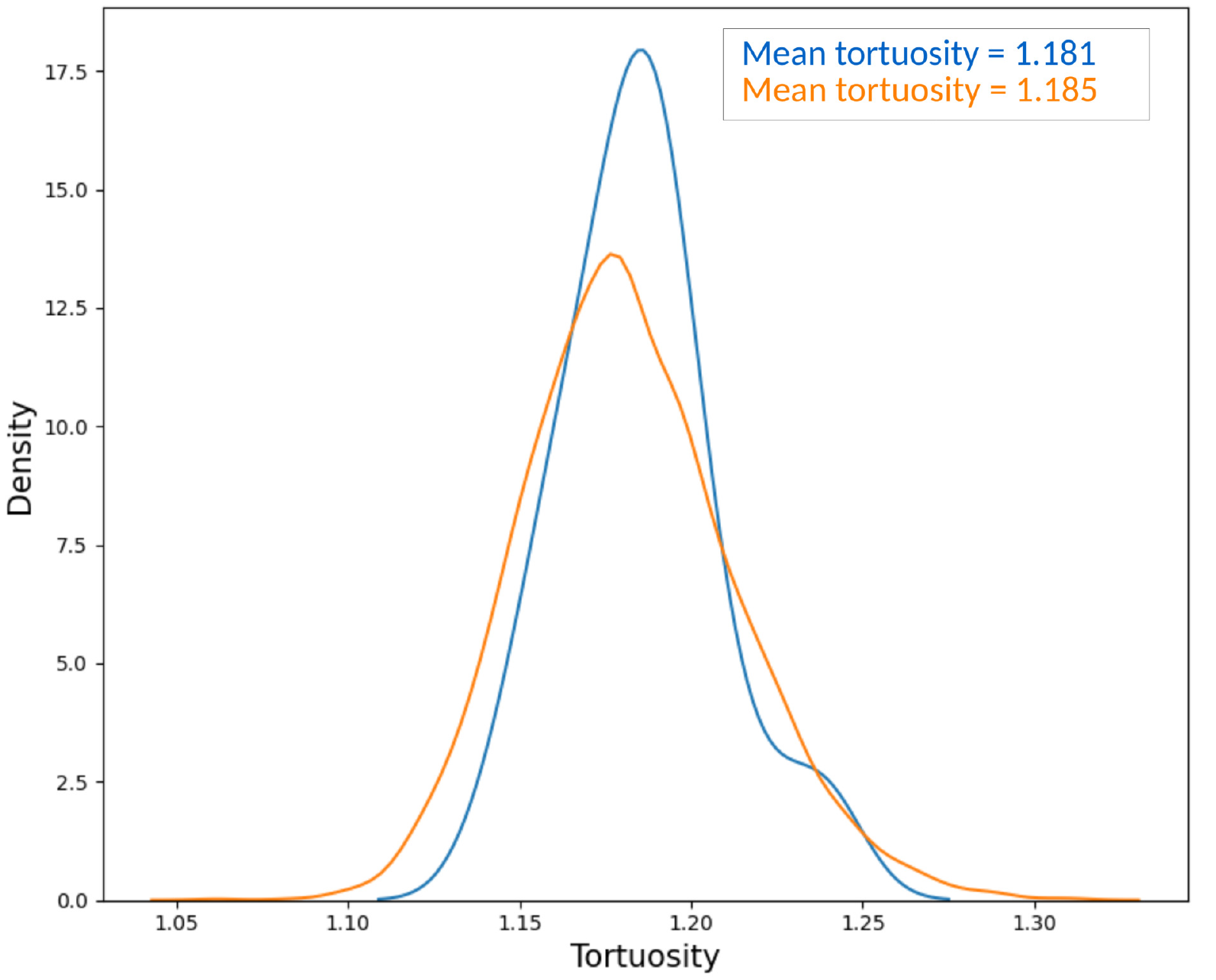}
 \caption{\label{fig:dens_mult} Comparison of the tortuosity densities constructed from the XPER simulations (blue) and the model predictions (orange) for microstructures with multiform aggregates (density obtained by kernel density estimate method)}
\end{figure}

Finally, the performance of the model is studied in term of tortuosity density estimate taking into account the variability due to all microstructures. \figurename~\ref{fig:dens_mult} shows that the XPER tortuosity density can be accurately approximated. 

\newpage
\section{Conclusion}
We develloped a probabilistic model for 2D fast crack prediction. This model leads to an algorithm that approximate at a low computational cost the local path of a crack. \\

The algorithm has been fully described and includes two steps. The first step is the computation of two local geometrical indicators (distance and angle) to capture the aggregate configurations in the vicinity of the crack tip. An efficient procedure exploiting a shadow cone has been proposed to evaluate them. The second step is the realization of a Markov chain model on the basis of the two indicators. The model parameters are estimated from a training set of numerically cracked microstructures. From any point of a crack in a microstructure, the model is able to evaluate the capability of any new point to be reached by the crack.  \\

The numerical results show that, for a given microstructure with different aggregate shapes, the new model provides a set of predicted cracks that are in agreement with the computer code simulation. Moreover, it allows recovering the variability of mechanical quantities of interest such as the tortuosity.  \\

The main advantage for practical issues is that the model is sufficiently flexible to be automatically adapted to any change in the local direction of the crack. A more complex application (a three-point bending beam test) is currently conducted on a mechanical test case where the crack direction is obtained through a mechanical simulation. \\

In this paper, it was decided to study microstructures of similar uniform density with different shapes of aggregates. 
Further investigations should concern different microstructures, in particular non-uniform densities (presence of clusters) and aggregates of different sizes.
The first step will be to test the robustness of the current model on these new types of microstructures. According to the results, adaptation of the model should be considered, taking in mind that our goal is not to calculate a new training set for each new case.

\newpage
\printbibliography

\end{document}